 \documentclass[pmlr,twocolumn,10pt]{jmlr} 

\usepackage{booktabs}
\usepackage{graphicx}
\usepackage{subcaption}
\usepackage{multirow}

\usepackage{algorithm}      
\usepackage{algpseudocode}  
\usepackage{multicol}       
\usepackage{amsmath}        

\usepackage{siunitx}

\usepackage[switch]{lineno}

\newcommand{\equal}[1]{{\hypersetup{linkcolor=black}\thanks{#1}}}

\theorembodyfont{\upshape}
\theoremheaderfont{\scshape}
\theorempostheader{:}
\theoremsep{\newline}

 \title[Inferring Optical Tissue Properties using Hybrid Amortized Inference]{Inferring Optical Tissue Properties from Photoplethysmography using Hybrid Amortized Inference}

\author{
 \Name{Jens Behrmann}\equal{Equal contribution, ordered alphabetically.}\footnotemark[2]\footnotemark[4],
 \Name{Maria R. Cervera}\footnotemark[1]\footnotemark[2]\footnotemark[4],
 \Name{Antoine Wehenkel}\footnotemark[1]\footnotemark[2]\footnotemark[4],
 \Name{Andrew~C. Miller}\footnotemark[2],
 \Name{Albert Cerussi}\footnotemark[2],
 \Name{Pranay Jain}\footnotemark[2],
 \Name{Vivek Venugopal}\footnotemark[2],
 \Name{Shijie Yan}\footnotemark[2],
 \Name{Guillermo Sapiro}\footnotemark[2],
 \Name{Luca Pegolotti}\footnotemark[3],
 \Name{J\"orn-Henrik Jacobsen}\footnotemark[3]
}

\begin{document}

\maketitle

\makeatletter
{ 
\renewcommand{\thefootnote}{\fnsymbol{footnote}}
\renewcommand{\@makefntext}[1]{\noindent\makebox[1.8em][r]{\@thefnmark\,}#1}
\footnotetext[2]{Apple.}
\footnotetext[3]{Work done while at Apple.}
\footnotetext[4]{\texttt{\{j\_behrmann, m\_cervera, awehenkel\}@apple.com}
}
\makeatother

\begin{abstract}
Smart wearables enable continuous tracking of established biomarkers such as heart rate, heart rate variability, and blood oxygen saturation via photoplethysmography (PPG). Beyond these metrics, PPG waveforms contain richer physiological information, as recent deep learning (DL) studies demonstrate. However, DL models often rely on features with unclear physiological meaning, creating a tension between predictive power, clinical interpretability, and sensor design.
We address this gap by introducing PPGen, a biophysical model that relates PPG signals to interpretable physiological and optical parameters. Building on PPGen, we propose hybrid amortized inference (HAI), enabling fast, robust, and scalable estimation of relevant physiological parameters from PPG signals while correcting for model misspecification. In extensive in-silico experiments, we show that HAI can accurately infer physiological parameters under diverse noise and sensor conditions.
Our results illustrate a path toward PPG models that retain the fidelity needed for DL-based features while supporting clinical interpretation and informed hardware design.

\end{abstract}

\paragraph*{Data and Code Availability}
We use in-silico datasets generated by the proposed biophysical model (PPGen). Generating this data relies on an internally developed light transport software, which prevents us from sharing the data. Due to our inability to share the in-silico dataset, we do not make our code available.

\section{Introduction}\label{sec:intro}
Wearable photoplethysmogram (PPG) sensors—small devices that emit light into tissue and measure the transmitted or reflected light—have become central to personal health and fitness monitoring \citep{Charlton_2023, kyriacou2021photoplethysmography}. From the outset, they have enabled the extraction of clinically meaningful biomarkers such as blood oxygen saturation, heart rate, and heart rate variability, all of which rest on well-understood biophysical principles. 

More recently, deep learning (DL) has reshaped how PPG signals are analyzed. Beyond traditional biomarkers, DL models have shown that PPG waveform dynamics contain information useful for detecting a wide range of medical conditions \citep{abbaspourazad2024largescale}. While these advances highlight the rich information encoded in PPG signals, DL models often exploit features that lack clear physiological interpretation. This lack of interpretability not only complicates clinical adoption but also hinders the design of next-generation PPG sensors, which must guarantee that such features are captured with sufficient fidelity to be reliably useful.

To align our theoretical understanding of PPG with the potential of DL-based predictive models, we propose a hybrid modeling approach that combines first-principles PPG models with deep generative models. This approach is designed to infer underlying physiological parameters while remaining robust to model misspecification, thereby overcoming the limitations of first-principles models without sacrificing predictive power or interpretability. 

Our contributions are threefold. \textbf{1.} We introduce PPGen, a novel PPG Pulse Generator that explicitly relates waveform characteristics to both physiological parameters and sensor architecture, using physically grounded light-transport simulations. \textbf{2.} We develop hybrid amortized inference (HAI), a learning algorithm that enables fast, robust, and scalable parameter inference while mitigating the impact of model misspecification. \textbf{3.} We systematically evaluate the model and inference framework in-silico across diverse configurations, demonstrating their ability to accurately estimate relevant physiological parameters from PPG measurements.

\begin{figure*}[ht]
    \centering
    \includegraphics[width=1.0\textwidth]{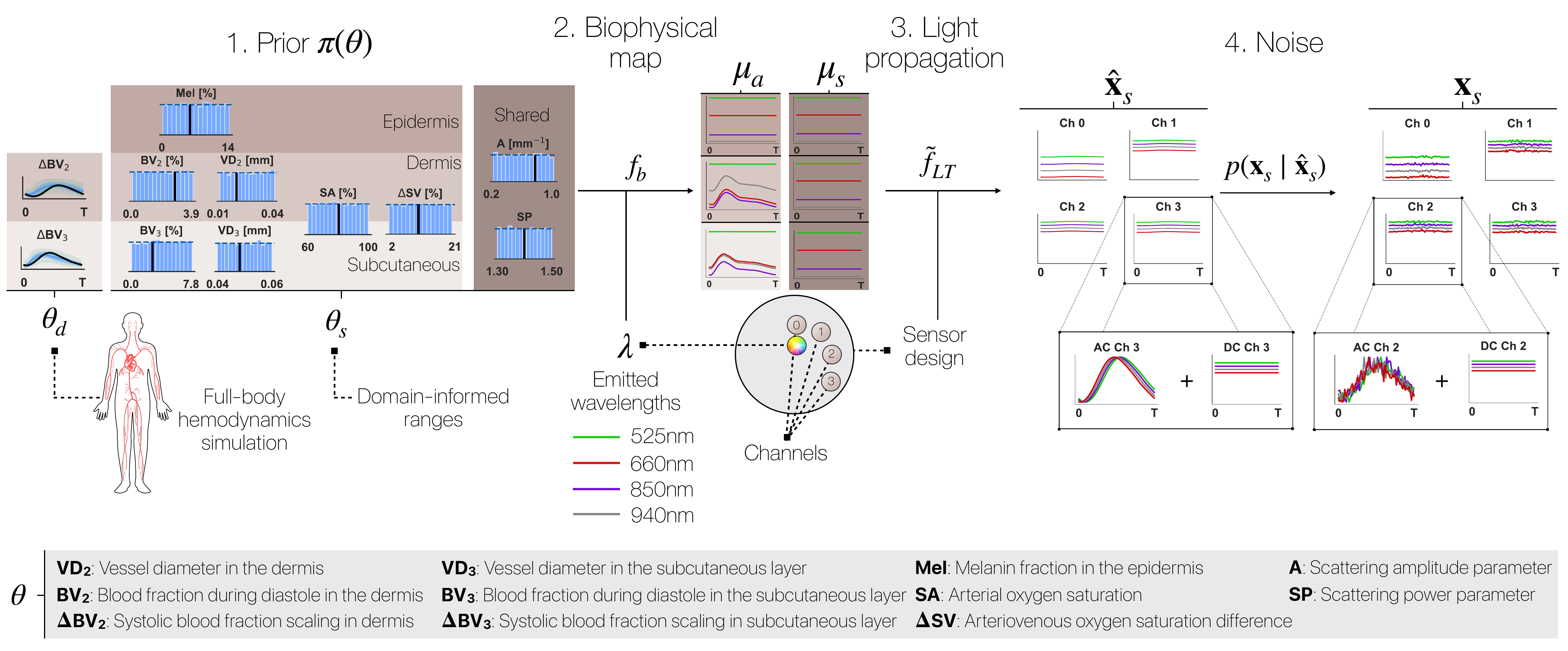}
    \caption{\small Overview of PPGen, our biophysical forward model for generating synthetic PPG signals. \textbf{1.} The process starts by sampling a set of biophysical parameters $\theta$ from a prior distribution $\pi(\theta)$. This prior combines dynamic parameters ($\theta_d$), such as blood volume waveforms from a hemodynamics simulator, with static parameters ($\theta_s$), such as melanin fraction and vessel diameters, sampled from literature-informed ranges. \textbf{2.} These parameters, along with a specific light wavelength ($\lambda$), are given as inputs to a biophysical mapping function, $f_b$, which calculates the optical absorption ($\mu_a$) and scattering ($\mu_s$) coefficients for each skin layer. \textbf{3.} Subsequently, a light transport model, $\hat{f}_{LT}$ (a neural network surrogate for Monte Carlo simulations), uses these optical coefficients and a given sensor architecture to predict the clean, noiseless PPG signal, $\hat{\mathbf{x}}_s)$. \textbf{4.} In the final step, a realistic noise model, $p(\mathbf{x}_s \mid \hat{\mathbf{x}}_s)$, adds shot and electronic noise to produce the final synthetic raw sensor reading, $\mathbf{x}_s$. \autoref{alg:PPGen} in \autoref{app:PPGen} describes sampling and density evaluation. Appendix \autoref{app:PPGen} further details each building block of PPGen.
    }
    \label{fig:overview_pipeline}
    \vspace{-1em}
\end{figure*}
\section{Methods}
\label{sec:methods}

We restrict temporal signals to single PPG pulses, i.e. characteristic waveforms representing the cyclical change in blood volume in a peripheral artery with each heartbeat.
We denote an \textit{observed} real-world PPG pulse as $\mathbf{x}_o \in \mathbb{R}^{R \times N \times T}$, where $R, N, T$ respectively denote the number of light receivers, light sources, and timesteps in the considered pulse. 

From a given PPG pulse $\mathbf{x}_o$, our goal is to infer a set of underlying biophysical parameters of interest $\theta \in \mathbb{R}^{9 + 2T}$~(see  appendix \autoref{tab:PSRs} for details), including $9$ static parameters (e.g. melanin concentration or arterial oxygenation) and two dynamic parameters that represent blood volume changes over the course of a PPG pulse. Formally, our objective is to learn an estimator that describes the posterior distribution of parameters $\theta$ given any PPG signal $\mathbf{x}_o$, i.e., $p(\theta \mid \mathbf{x}_o)$.  

Since in-vivo measurement of biophysical tissue properties $\theta$ is invasive, we cannot rely on labeled data to solve this task. To mitigate this problem, we distill expert knowledge into a model $\pi(\theta)p(\mathbf{x}_s \mid \theta)$ that describes a joint generative process of biophysical parameters $\theta$ and \textit{synthetic} PPG pulses $\mathbf{x}_s \in \mathbb{R}^{R \times N \times T}$. This model provides a good approximation of the \textit{true} relationship $p(\mathbf{x}_o \mid \theta)$. However, some of its simplifying assumptions, e.g., ignoring movement artifacts and assuming a fixed geometry of skin layers, reduce its ability to accurately describe real-world PPG pulses $\mathbf{x}_o$. 

In summary, our goal of estimating $p(\theta \mid \mathbf{x}_o)$ is challenged by \textbf{1.} Data scarcity: we only have unlabeled real-world data $\mathcal{D} := \{\mathbf{x}_o\}_{i=1}^N$; and \textbf{2.} Model misspecification: because the underlying mechanisms affecting PPG measurements are numerous and complex, first-principle models are imperfect. In \autoref{sec:model}, we introduce PPG Pulse Generator~(PPGen), a differentiable biophysical model $p(\mathbf{x}_s \mid \theta)\pi(\theta)$ that helps us address data scarcity.

With PPGen defined, the main methodological obstacle remaining for estimating $p(\theta \mid \mathbf{x}_o)$ is the misspecification gap between $\mathbf{x}_s$ and $\mathbf{x}_o$. To address it, we introduce in \autoref{subsec:hvae-inference} hybrid amortized inference~(HAI), our solution that learns jointly from unlabeled real-world data $\mathcal{D}$ and from labeled synthetic measurements $(\theta, \mathbf{x}_s) \sim  p(\mathbf{x}_s \mid \theta) \pi(\theta)$. In \autoref{subsec:implementation} we discuss the  main modeling choices we followed to conduct our experiments.

\subsection{PPG Pulse Generator~(PPGen)}\label{sec:model}
We model skin as a multi-layered material composed of $l=3$ layers. We can then parameterize the optical skin properties with some biophysical parameters $\theta \in \Theta$ that describe the main biophysical factors impacting the propagation of light in skin. In particular, these parameters can be decomposed into $\theta =  [\theta_s, \theta_d ]$. The static parameters, denoted $\theta_s \in \mathbb{R}^{9}$, are fixed over the measurement period considered, whereas the dynamic parameters $\theta_d \in \mathbb{R}^{2T}$ drive the time-dependence of the PPG signal.
The likelihood can thus be decomposed into $p(\mathbf{x}_s \mid \theta) = \prod_{t=1}^T p(\mathbf{x}_s^t \mid \theta_s, \theta_d^t),$ where $\theta_d^t \in \mathbb{R}^2$ are the two dynamic parameters at timestep $t$ and $\mathbf{x}_s^t \in \mathbb{R}^{R \times N}$. For readability, we denote $\theta^t \in  \mathbb{R}^{11}$ the vector composed of $\theta_s$ and $\theta_d^t$.

We decompose the time-independent joint distribution $p(\theta^t, \mathbf{x}_s^t)$ into four main components: \textbf{1.} an informed prior $\pi(\theta)$ over the biophysical parameters; \textbf{2.} a function $f_b$ that maps the instantaneous parameters $\theta^t$ to two vectors $\mu_a \in \mathbb{R}^{l}$ and $\mu_s \in \mathbb{R}^{l}$ describing light absorption and scattering in the $l$ considered skin layers; \textbf{3.} a neural network based surrogate of a light propagation simulator in layered materials, denoted $\hat{f}_{\text{LT}}$, which maps optical properties $(\mu_a, \mu_s)$ to the instantaneous PPG the sensor would measure were it noiseless; and \textbf{4.} a noise model of the sensor.

\paragraph{$\pi(\theta)$ -- Informed prior over biophysical parameters.} For the set of static parameters $\theta_s$, we sample from a uniform distribution over a range of plausible values obtained from the literature. For the two dynamic parameters $\theta_d$ consisting of blood volume changes over time, we leverage mechanistic models of the cardiovascular system~\citep{Charlton2019, melis2018openbf} and sample over plausible ranges of cardiac and vascular properties to obtain a variety of blood volume waveforms. For additional details on $\theta \sim \pi(\theta)$ please refer to \autoref{app:prior}.

\paragraph{$f_b$ -- Modeling the relationship between biophysical parameters and optical skin properties.}
The function  $f_b(\theta^t, \lambda): \Theta^t \times \Lambda \rightarrow \mathbb{R}^l \times \mathbb{R}^l$ relates biophysical parameters $\theta$ to the tissue's optical properties, $\mu_a\in \mathbb{R}^{l}$ and $ \mu_s \in \mathbb{R}^{l}$, for a given wavelength $\lambda \in \Lambda$, where the $l=3$ skin layers correspond to the epidermis, dermis, and subcutaneous.
The absorption coefficient $\mu_a$ is a function of the concentration of absorbing components in skin, such as hemoglobin or melanin (the complete list is described in \autoref{fig:overview_pipeline}). By contrast, $\mu_s$ describes the trajectories that light rays follow in skin, and is described in our model by the scattering parameters A and SP.

Crucially, because both of these properties depend on wavelength $\lambda$, $f_b$ takes $\lambda$ as an additional input. For additional details on how $\mu_a$ and $\mu_s$ are computed, please refer to \autoref{app:PSR_map}.

\paragraph{$\hat{f}_{\text{LT}}$ -- Modeling light propagation in multi-layered tissue via an efficient and differentiable surrogate.}
The propagation of light within layered materials is well understood and can be simulated with GPU-accelerated Monte Carlo modeling of light transport in multi-layered tissues~\citep[MCML,][]{wang1995mcml, Alerstam2008}.Simulations depend on the layer-dependent optical tissue properties $\mu_a\in \mathbb{R}^l$ and $\mu_s \in \mathbb{R}^l$, the sensor architecture (e.g., the emitter and detector locations) and the layers' geometry (e.g., layer thicknesses). The two latter are considered fixed in our experiments, unless stated otherwise. For a given pair of optical parameters $(\mu_a, \mu_s)$, the output of the MCML simulation $f_{\text{LT}}(\mu_a, \mu_s) \in \mathbb{R}^{R}$ is an estimator of detected versus emitted light at each of the $R$ light detectors. In our experiments, we use a light transport tool that was extensively validated against MCML and consider sensors with four receiving photo-diodes and one shared emitter location, hence $R = 4$.

While faithful to the physics, light transport simulators are computationally intensive and pose a major bottleneck in the generative process. We address this limitation by replacing $f_{\text{LT}}$ with a fast neural network, $\hat{f}_{\text{LT}}$, that is trained via supervised learning on a large-scale dataset of optical-property-to-measurement pairs, $\left\{\left(\mu_a^{(i)}, \mu_s^{(i)} \right), \mathbf{x}_s^{(i)}\right\}_{i=1}^{N_{sim}}$, obtained by running the light transport simulator on a wide range of inputs (for details refer to \autoref{app:surrogate_training}). The resulting light transport simulator surrogate $\hat{f}_{\text{LT}} \approx f_{\text{LT}}$ can generate instantaneous and noiseless PPG signals $\hat{\mathbf{x}}_s^t \in \mathbb{R}^{R \times N}$ via $\hat{\mathbf{x}}_s^t  =\hat{f}_{\text{LT}}(\mu_a, \mu_s)$.
Because this surrogate has the added benefit of being differentiable, combining it with the analytical biophysical map $f_b$ results in a fully differentiable forward model $f_t(\theta^t) := \left[\hat{f}_{\text{LT}} \circ  f_b(\theta^t, \lambda_1), \dots, \hat{f}_{\text{LT}} \circ f_b(\theta^t, \lambda_N))\right]$, which concatenates along the wavelength dimension. To describe a complete pulse, we concatenate outputs of $f_t$ along time $t$ such that $f: \Theta \mapsto \mathbb{R}^{R \times N \times T}$. 

\paragraph{Noise sensor model.} \label{sec:noise}
We further improve the faithfulness of the model by adding measurement noise in the form of two additive, time-independent noise sources: \textit{shot noise}--a zero-mean Gaussian noise with signal-dependent variance--, and \textit{white noise}. While shot noise arises from the particle nature of light and models the random arrival of photons at the receiver, white noise captures sensor-internal effects, such as thermal and chip noise.
We denote this combined noise model as
 $p(\mathbf{x}_s \mid \hat{\mathbf{x}}_s)$ and provide details in \autoref{app:sensor_and_noise_model}.

\autoref{fig:overview_pipeline} depicts the whole PPGen, which provides a differentiable way of generating PPG pulses and evaluating the corresponding likelihood. 

\subsection{Hybrid Amortized Inference (HAI)}
\label{subsec:hvae-inference}

\begin{algorithm}[h]
\caption{Training and inference with Hybrid Amortized Inference~(HAI).}\label{algo:all}
\DontPrintSemicolon

\SetKwFunction{PretrainNPE}{PretrainNPE}
\SetKwFunction{LearnMisspec}{LearnMisspec}
\SetKwFunction{Infer}{Infer}
\BlankLine

\PretrainNPE{$\pi(\theta),\, p(\mathbf{x}_s\mid\theta)$}\;\\
\Repeat{convergence}{
  Sample $\theta \sim \pi(\theta),\; \mathbf{x}_s \sim p(\mathbf{x}_s\mid\theta)$\;\\
  $\mathcal{L}_s(\phi) \gets -\log q_\phi(\theta \mid \mathbf{x}_s)$\;\\
  $\phi \gets \phi - \eta \nabla_\phi \mathcal{L}_s(\phi)$\;
}
\Return $q_{\phi^\star}(\theta \mid \mathbf{x}_s)$\;

\BlankLine
\BlankLine
\LearnMisspec{$q_{\phi^\star},\, \mathcal{D}=\{\mathbf{x}_o\}$}\;\\
\Repeat{convergence}{
  \ForEach{$\mathbf{x}_o \in$ minibatch $\mathcal{D}$}{
    $\mathbf{x}_s \sim q_\psi(\mathbf{x}_s\mid \mathbf{x}_o)$\;\\
    $\theta \sim q_{\phi^\star}(\theta \mid \mathbf{x}_s)$\;\\
    $J(\psi,\omega) \gets 
       \log q_\omega(\mathbf{x}_o \mid \mathbf{x}_s) 
       + \log p(\mathbf{x}_s \mid \theta)$\;\\
    $(\psi,\omega) \gets 
       (\psi,\omega) + \eta \nabla_{(\psi,\omega)} J(\psi,\omega)$\;
  }
}
\Return $q_{\psi^\star}(\mathbf{x}_s\mid \mathbf{x}_o)$\;

\BlankLine
\BlankLine
\Infer{$\mathbf{x}_o, N_{\text{samples}}$}\;\\{
\For{$i=1,\dots,N_{\text{samples}}$}{
  $\mathbf{x}_s^{(i)} \sim q_{\psi^\star}(\mathbf{x}_s\mid \mathbf{x}_o)$\;\\
  $\theta^{(i)} \sim q_{\phi^\star}(\theta \mid \mathbf{x}_s^{(i)})$\;\\
}
\Return $\{\theta^{(i)}\}_{i=1}^{N_{\text{samples}}}$\;
}
\end{algorithm}

This section introduces Hybrid Amortized Inference (HAI), our algorithm to obtain an amortized estimator of the posterior $p(\theta \mid \mathbf{x}_o)$, given the prescribed model $p(\theta, \mathbf{x}_s)$ and unsupervised data $\mathcal{D}= \{\mathbf{x}_o^i\}_{i=1}^M$.

\paragraph{Misspecification model.} Because labels $\theta$ are not available for real measurements $\mathbf{x}_o$, we aim to learn a projection from $\mathbf{x}_o$ into a physical signal $\mathbf{x}_s$, from which parameters $\theta$ can be inferred. This projection, a misspecification model that corrects the gap between real and synthetic data, therefore needs to be independent of the real parameter $\theta$. This results in the independence assumption 
\begin{align}
\mathbf{x}_o \perp \theta \mid \mathbf{x}_s, \label{eq:assumption}
\end{align}
which essentially states that a real-world measurement $\mathbf{x}_o$ contains no more information about $\theta$ than the corresponding synthetic $\mathbf{x}_s$. As leveraged in previous work \citep{wehenkel2024addressing}, this assumption has the attractive property of resulting in the following decomposition: 
\begin{align}
    p(\theta \mid \mathbf{x}_o) = \int p(\theta \mid \mathbf{x}_s) p(\mathbf{x}_s \mid \mathbf{x}_o) \text{d}\mathbf{x}_s.\label{eq:factor_misspecification}
\end{align}
Our original objective of estimating $p(\theta \mid \mathbf{x}_o)$ can thus be disentangled into learning independent estimators for $p(\theta \mid \mathbf{x}_s)$ and $p(\mathbf{x}_s \mid \mathbf{x}_o)$. 

\paragraph{Neural posterior estimation of $p(\theta \mid \mathbf{x}_s)$.}
Following the neural posterior estimation~(NPE) algorithm introduced by \citet{lueckmann2017flexible}, we train a neural conditional density estimator on pairs of parameters and simulated observations $(\theta, \mathbf{x}_s) \sim  p(\mathbf{x}_s \mid \theta) \pi(\theta)$ directly. Formally, we consider a family of approximators parameterized by $\phi$ denoted, $q_\phi(\theta \mid \mathbf{x}_s)$, and perform stochastic gradient descent on $\phi$ to minimize
\begin{align}
    \mathcal{L}_s(\phi) &:= -\frac{1}{L}\sum_{i=1}^L \log q_{\phi}(\theta^i \mid \mathbf{x}_s^i) \\ 
    & \text{with} \; (\theta^i, \mathbf{x}^i_s) \sim p(\mathbf{x}_s \mid \theta) \pi(\theta)\nonumber.
\end{align}
Provided sufficient capacity and proper optimization, the model obtained, denoted $q_{\phi^\star}(\theta \mid \mathbf{x}_s)$, can be made arbitrarily close to the ``true'' posterior $p(\theta \mid \mathbf{x}_s)$ on synthetic data.

\paragraph{Maximum likelihood estimation of the misspecification model $p(\mathbf{x}_s \mid \mathbf{x}_o)$.}
We consider a family of misspecification approximators $q_{\psi}(\mathbf{x}_s \mid \mathbf{x}_o)$, parameterized by a vector $\psi \in \mathbb{R}^{\lvert \psi \rvert}$. Then, the likelihood of the parameter $\psi$ given a dataset of observation $\mathcal{D}$ becomes
\begin{align}
    &\log p(\mathcal{D}\mid \psi) =  \sum_{\mathbf{x}_o \in \mathcal{D}}  \log p(\mathbf{x}_o \mid \psi) \label{eq:ELBO}\\
    &\approx  \sum_{\mathbf{x}_o \in \mathcal{D}}  \mathbb{E}_{\substack{q_{\psi}(\mathbf{x}_s \mid \mathbf{x}_o) \\[2pt] 
                      q_{\theta^\star}(\theta \mid \mathbf{x}_s)}}[ q_{\omega}(\mathbf{x}_o \mid \mathbf{x}_s) p(\mathbf{x}_s \mid \theta)], \label{eq:MLE_miss}
\end{align}
where $q_{\omega}(\mathbf{x}_o \mid \mathbf{x}_s)$ approximates $p(\mathbf{x}_o \mid \mathbf{x}_s) = \frac{q_{\psi}(\mathbf{x}_s \mid \mathbf{x}_o) p(\mathbf{x}_o)}{p(\mathbf{x}_s)}$, the inverse of  $q_{\psi}(\mathbf{x}_s \mid \mathbf{x}_o)$. We can directly convert \autoref{eq:MLE_miss} into a loss function if the approximators $q_{\omega}(\mathbf{x}_o \mid \mathbf{x}_s)$ and $q_{\theta^\star}(\theta \mid \mathbf{x}_s)$ allow differentiable sampling, and if $q_{\psi}(\mathbf{x}_s \mid \mathbf{x}_o)$ allows differentiable density evaluation.

In summary, HAI combines NPE with maximum likelihood estimation to learn a model of the misspecification, while leveraging the independence assumption in \autoref{eq:factor_misspecification}. The training and inference algorithms are described in \autoref{algo:all} (more details in \autoref{app:algs}).

\subsection{Variational families and implementation details}\label{subsec:implementation}
As highlighted in the previous section, the HAI framework heavily relies on differentiable parameterization of conditional probability distributions. We now provide details on the neural network architectures used to model these various functions.

\paragraph{Parameterization of $q_\phi$.}
In our experiments we use a simple variational family that assumes conditional independence between all elements of $\theta$ given the observation $\mathbf{x}_s$ so that we can re-express the posterior as $q_\phi(\theta \mid \mathbf{x}_s) = \Pi_{i=1}^{9 + 2T} q^i_\phi(\theta_i \mid \mathbf{x}_s)$. The parametric density for $q^i_\phi$ is a fixed-variance Gaussian whose mean is determined by the output of a neural network parameterized by $\phi$,  $q^i_\phi(\theta_i \mid \mathbf{x}_s) := \mathcal{N}(f^i_{\phi}(\mathbf{x}_s) - \mu_i, \sigma_i^2)$, where $\mu_i$ and $\sigma_i$ respectively denote the mean and variance of the chosen prior for $\theta_i$. 
In our experiments, we use the posterior mean as the point estimate for $\theta$. We use a shared U-net architecture to jointly model $f^i_{\phi}(\mathbf{x}_s)$ for all $i \in [1, 9+ 2T]$. Importantly, we also engineer the input to the U-net and explicitly decompose the signal into a pulsatile (AC) and a quasi-static (DC) baseline components, which have typically very different scales that would be hard for the neural network to pick up. More details about the neural network architecture are provided in \autoref{app:ppg_features}. Alternatives to this simple model include conditional normalizing flows, which would enable to model uncertainty on $\theta$ jointly and highlight dependencies among parameters. 

\paragraph{Misspecification model $q_{\psi}(\mathbf{x}_s \mid \mathbf{x}_o)$.}
To demonstrate the importance of carefully designing the inductive bias of $q_{\psi}(\mathbf{x}_s \mid \mathbf{x}_o)$, our experiments rely on a simple additive correction model, $q_{\psi}(\mathbf{x}_s \mid \mathbf{x}_o, \theta) := \delta_{\mathbf{x}_o}(\mathbf{x}_s + \beta(\psi))$, where $ \delta$ denotes the Dirac delta distribution and $\beta: \mathbb{R}^{R\times N } \rightarrow \mathbb{R}^{R\times N \times T}$ denote an offset shared over time and computed from $\psi \in \mathbb{R}^{R \times N}$. A perfect estimator of the inverse misspecification $p(\mathbf{x}_o \mid \mathbf{x}_s) = \frac{q_{\psi}(\mathbf{x}_s \mid \mathbf{x}_o) p(\mathbf{x}_o)}{p(\mathbf{x}_s)}$ can be derived analytically as $q_{\psi}(\mathbf{x}_o \mid \mathbf{x}_s) := \delta_{\mathbf{x}_s}(\mathbf{x}_o  - \beta(\psi))$. In practice, one could consider more expressive misspecification models, e.g., parameterized with a neural network. However, this may require careful regularization to ensure the learnt misspecification model does not overule the prescribed model $p(\mathbf{x}_s \mid \theta)$. Our experiments will further discuss the trade-offs between expressivity and generalization of the learned misspecification model.

\section{Related work}

\paragraph{Mechanistic modeling of PPG.} 
Similar to our work, both \citet{boonya2020modeling} and \citet{Tang2020synthetic} capture the dynamic nature of PPG using light transport simulations. While \citet{boonya2020modeling} introduce a dynamic 3D model of blood vessels, \citet{Tang2020synthetic} approximate PPG pulses with waveform templates. In contrast, our model leverages the full-body hemodynamics simulator of \citet{Charlton2019} to describe pulse dynamics. It further incorporates Windkessel models~\citep{Frasch1996} to capture microvascular effects, as in \citet{Tanaka_2022} for PPG waveform modeling or \citet{Doostdar2014quantification} for studying aging.
Our model combines these hemodynamic waveforms with Monte Carlo simulations of light propagation in multilayered tissue~\citep{jacques1995monte, wang1995mcml}. Recent work has has proposed important links between optical properties and skin physiology: e.g. \citet{Chatterjee2020investigating} investigated the origin of PPGs across multiple wavelengths; \citet{Haque2022} and \citet{Reiser2022} modeled the impact of blood vessels; and \citet{Al-Halawani2024} analyzed pigmentation via melanin absorption. Building on this, our simulator, PPGen, integrates mechanistic models of skin optics and pulsatile blood flow, enabling it to accurately generate PPG pulse waveforms while preserving the underlying links to sensor physics and physiology.

\paragraph{Inference of biophysical parameters on PPG.}

Existing methods for inferring tissue properties, whether simulation lookup tables (e.g. \cite{Zhong2014reflectanceInversion} and \cite{Das2021databaseinversion}) or recent machine learning models (see Table 1 in \cite{Scarbrough2024mlInversion} for an overview), limited their scope to inferring static optical parameters. Contrarily, our work extends the inference to biophysical parameters and dynamic waveforms crucial for PPG analysis. Furthermore, these two approaches have distinct shortcomings: classical methods rely on slow, iterative optimization that scales poorly \citep{Fredriksson2012inverse}, while modern ML methods use data augmentation for robustness \citep{Scarbrough2024mlInversion}, which fails against errors unseen during training. We address these gaps using hybrid amortized inference (HAI) which allows scalable, dynamic waveform estimation and robust adaptation to misspecification.

\paragraph{SBI/ Hybrid inference under misspecification.}
A key challenge in Simulation-Based Inference~\citep[][SBI]{cranmer2020frontier}, is its sensitivity to model misspecification \citep{cannon2022investigating}, a limitation that has spurred the development of robust methods \citep{wehenkel2024addressing, ward2022robust, huang2023learning, kelly2023misspecification}. Among these, HAI is closely related to RoPE \citep{wehenkel2024addressing}, as both leverage the independence assumption from \autoref{eq:assumption} to improve robustness. However, unlike RoPE, HAI requires a differentiable forward model to explicitly learn a misspecification model. This explicit learning of misspecification connects HAI to Generalized Bayesian Inference \citep[][GBI]{bissiri2016general, jewson2018principled, cherief2020mmd, matsubara2022robust, dellaporta2022robust}, which replaces the initial likelihood with a generalized loss-based likelihood. The primary novelty of HAI compared to previous GBI approaches is its use of a generative model to learn the misspecification, allowing it to benefit from advances in deep generative modeling. HAI is further connected to physics-integrated autoencoders \citep{takeishi2021physics}, which offer a hybrid learning strategy for inference under misspecification. Compared to this generic approach, HAI simplifies the training procedure into two distinct steps using the independence assumption from \autoref{eq:assumption}.

\section{Results}

Through a series of in-silico experiments, we now demonstrate the potential of combining PPGen and HAI to infer interpretable biophysical parameters from real-world PPG measurements.

\begin{figure}[t]
    \centering
    \includegraphics[width=\linewidth]{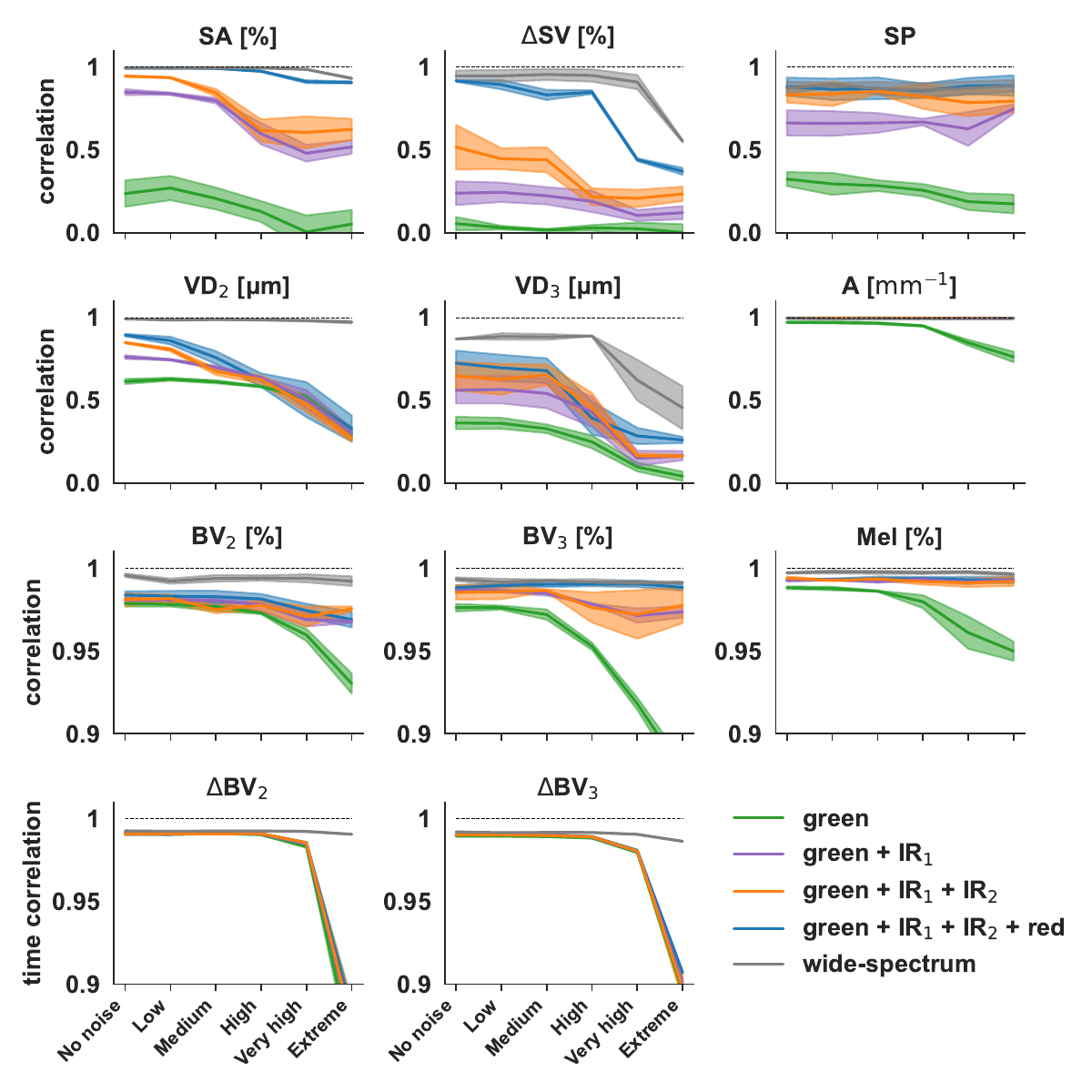}
    \caption{\small  Identifiability analysis of biophysical parameters under varying noise levels and wavelength configurations. We report the Pearson correlation coefficient between the groundtruths and the values inferred from $\mathbf{x}_s$. For dynamic properties, we report averages of across-time correlation coefficients. We use three random seeds to estimate the variation across different training runs.}
    \label{fig:clean_results}
\end{figure}

\subsection{Inference of biophysical properties under multiple PPG wavelength configurations}
\label{results:noise}
In \autoref{fig:clean_results}, we analyze the identifiability of the biophysical parameters $\theta$ as a function of emitter wavelengths and noise levels in $\mathbf{x}_s$. The noise levels, shown in \autoref{fig:noise_viz}, range from optimistic to pessimistic hardware noise specifications. The wavelength configurations (subsets of green, infrared, red) mimic standard sensor configurations and are compared to a wide-spectrum setup covering $500$–$1000$nm at $1$nm resolution (see \autoref{app:sensor_and_noise_model} for details on sensor and noise settings). For this analysis, we use only the NPE model $q_\phi$ described in \autoref{sec:methods}, trained on matching simulations (details in \autoref{app:exp_details}).

Focusing on the $9$ static parameters in \autoref{fig:clean_results}, we find that the wide-spectrum sensor consistently outperforms all other configurations across noise levels and parameters (see \autoref{fig:zoom_in_static} for a breakdown of results at a medium level of noise). Overall these results show the ability of NPE to learn an effective inference procedure, where identifiability slowly declines with increasing noise levels. Focusing on common sensor configurations with few wavelengths we observe that some parameters, particularly those modeling vessel diameters, are challenging to estimate even under optimal conditions. Interestingly, we observe that oxygen saturation parameters (SA, $\Delta$SV) are only robustly identified by multi-wavelength sensors. This well-known biophysical fact~\citep{tamura2019current} provides evidence that PPGen faithfully represents some underlying causes of PPGs. 

Analyzing the identifiability of dynamic parameters ($\Delta\text{BV}_2$, $\Delta\text{BV}_3$) in \autoref{fig:clean_results}, we observe high reconstruction quality across wavelength configurations up to reasonably high noise levels. As shown in \autoref{fig:bv_inferences}, no reconstruction falls below $98\%$ time-correlation under medium noise. This interesting in-silico result suggests that even green-only sensors can disentangle a PPG waveform into two blood volume components and improve reconstruction quality, likely helped by strong priors on waveform structure. Since these inferred waveforms can be mechanistically linked to fundamental cardiovascular parameters (see \autoref{app:blood_volume_sim}), they may open new directions for monitoring cardiovascular health.

\begin{figure}[t]
    \centering
    \includegraphics[width=\linewidth]{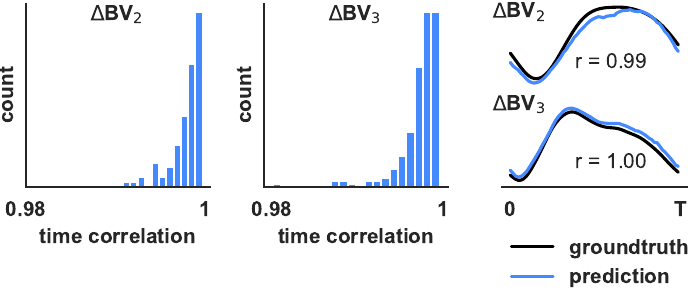}
    \caption{\small Inferences of blood volume waveforms with a four-wavelength PPG sensor without misspecification at medium noise level. Histograms show the distribution of across-time correlation coefficients between groundtruths and predictions, and the right hand side shows an example waveform pair.}
    \label{fig:bv_inferences}
    \vspace{-1em}
\end{figure}

\subsection{Robustness to misspecification}
\label{results:misspec}
In \autoref{tab:results_misspec}, we assess the ability of HAI to overcome model misspecification, which is expected to be present in real-world data. We consider the following misspecification settings: \textbf{None}: no model misspecification; \textbf{Noise}: forward model underestimates noise in observed data; \textbf{Sensor}: geometry of the sensor is imperfectly captured, which we model as constant per channel shift in observed data; \textbf{Skin}: imperfect modeling of skin, which we model by using a different skin layer thicknesses; \textbf{Combined}: combination of \textit{Noise}, \textit{Sensor}, and \textit{Skin} misspecification. A visualization of these effects on PPG signals is provided in \autoref{fig:miss_viz} and details are provided in \autoref{app:sensor_and_noise_model}. We compare HAI with two baselines. \textbf{Sim-only} directly applies NPE, without any learned correction and therefore lacks the ability to handle potential misspecification. \textbf{Real-only} optimizes jointly the encoder $q_\phi$ and misspecification $\psi$ to maximize \autoref{eq:MLE_miss}, and therefore lacks physical inductive biases introduced via NPE pretraining. Experimental details are provided in \autoref{app:exp_details}. 

\begin{table}[t]
  \caption{\small Parameter inference under several misspecifications, as measured by Pearson correlation and mean absolute percentage error~(MAPE) averaged over all biophysical parameters. For each result we report $\pm$ the standard deviation over multiple random seeds.}\label{tab:results_misspec}
  \vspace{-.5em}
  
  \setlength{\tabcolsep}{4pt}
  \begin{tabular}{llll}
  \toprule
  \bfseries Misspec. & \bfseries Method & \bfseries Correlation  & \bfseries MAPE \\
  \midrule
  
    \multicolumn{4}{c}{$4$-wavelength sensor}\\
    \cmidrule(lr){1-4}
      \multirow{ 3}{*}{None} & HAI & $0.93 \pm 0.01$ & $5.9 \pm 0.6$  \\
       & Sim-only & $0.93 \pm 0.01$ & $5.9 \pm 0.6$ \\
       & Real-only & $0.74 \pm 0.01$ & $14.6 \pm 0.6$ \\
       \cmidrule(lr){1-4}
    
    \multirow{ 3}{*}{Noise} & HAI & $\mathbf{0.83 \pm 0.01}$ & $\mathbf{10.5 \pm 0.4}$  \\
       & Sim-only & $0.81 \pm 0.01$ & $11.8 \pm 1.0$ \\
       & Real-only & $0.73 \pm 0.01$ & $16.9 \pm 0.6$ \\
       \cmidrule(lr){1-4}
    
    \multirow{ 3}{*}{Sensor} & HAI & $\mathbf{0.96 \pm 0.00}$ & $\mathbf{5.2 \pm 0.2}$  \\
       & Sim-only & $0.82 \pm 0.00$ & $11.9 \pm 0.7$ \\
       & Real-only & $0.72 \pm 0.01$ & $17.1 \pm 0.5$ \\
       \cmidrule(lr){1-4}
    
    \multirow{ 3}{*}{Skin} & HAI & $0.71 \pm 0.01$ & $18.2 \pm 0.6$  \\
       & Sim-only & $0.70 \pm 0.02$ & $18.0 \pm 0.8$ \\
       & Real-only & $0.67 \pm 0.02$ & $19.1 \pm 1.4$ \\
       \cmidrule(lr){1-4}
    
    \multirow{ 3}{*}{Combined} & HAI & $\mathbf{0.67 \pm 0.01}$ & $\mathbf{22.6 \pm 0.4}$  \\
       & Sim-only & $0.65 \pm 0.01$ & $24.1 \pm 0.8$ \\
       & Real-only & $0.65 \pm 0.00$ & $22.1 \pm 1.0$ \\
    \specialrule{1pt}{1pt}{1pt} 
    \multicolumn{4}{c}{wide-spectrum sensor}\\
    \cmidrule(lr){1-4}
    \multirow{3}{*}{\textit{Combined}
    } & HAI & $0.84 \pm 0.00$ & $12.6 \pm 0.3$  \\
       & Sim-only & $0.84 \pm 0.00$ & $11.2 \pm 1.4$ \\
       & Real-only & $0.71 \pm 0.00$ & $15.2 \pm 0.2$ \\
    
  \bottomrule
  \end{tabular}
\end{table}

We first observe a performance decline under nearly all types of misspecification compared to the well-specified case (\textit{None}). Across all settings, HAI outperforms or matches the two baselines, with its advantage being most pronounced under \textit{Sensor} misspecification—where the inductive bias of the misspecification model aligns best with the actual gap between real and synthetic data. These contrasted results show that HAI can mitigate model misspecification, while also underscoring the need for careful validation to identify effective misspecification models in practice. 

Finally, wide-spectrum sensor results in \autoref{tab:results_misspec} suggest that more informative sensors can improve parameter estimation even under misspecification (see also extended results in \autoref{supp:results:misspec}). A parameter-specific breakdown in \autoref{fig:misspec_watch_barplots} and \autoref{fig:misspec_wide_barplots} reveals two points: first, HAI consistently achieves the best estimates for arterial oxygen saturation, and second, the same parameters that were difficult to estimate under noisy settings are difficult to infer under misspecfication. 

\subsection{Scalability of HAI} %
\label{sec:scaling}
The results presented highlight the scalability of combining PPGen with HAI for real-world PPG analysis. Indeed, our framework handles wide-spectrum sensors that increase data volume per sample by over $100\times$, as illustrated in \autoref{fig:clean_results} and \autoref{tab:results_misspec}. These results also suggest that wide-spectrum sensors could serve as reference non-invasive devices to obtain approximate labels for biophysical parameters in human studies and help better understand the origins of PPG signals. Importantly, while training the neural networks ($\hat{f}_{LT}$, $q_\phi$, $q_\psi$) incurs a significant upfront cost, inference is nearly instantaneous and fully parallelizable on modern GPUs. In contrast, iterative inversion of the underlying light transport simulations requires repeated calls to the computationally expensive simulator, making large-scale analysis infeasible. By amortizing both simulations and inference, our framework enables the analysis of millions of PPG pulses efficiently, opening the door to population-level studies of biophysical parameter distributions.

\section{Discussion}
The clinical utility of PPG is often limited by a lack of annotations and the signal's poor interpretability.
In this paper, we address these challenges by adopting a hybrid ML philosophy. We first propose PPGen, an efficient PPG pulse simulator that is deeply rooted in biophysics and captures the entire generative process from cardiac parameters to PPG signals. 
Leveraging this efficient simulator, we then propose HAI, a framework for inferring biophysical parameters from PPG data. 
Our in-silico experiments show that HAI can robustly and scalably infer these parameters, even under noise and model misspecification.

A key advantage of our framework is its ability to enhance the interpretability of PPG signals. 
For example, while PPG waveforms are difficult to interpret, decomposing them into contributions from different parts of the peripheral vascular system (i.e. blood volume waveforms at different skin depths) opens the door to non-invasive monitoring of important cardiovascular parameters that are otherwise difficult to estimate.
This interpretability also extends to sensor design, where inferring tissue scattering properties can inform the development of next-generation devices that more precisely target specific skin structures.

Though grounded in established biophysical principles, PPGen only approximates the processes underlying real-world PPG measurements. For example, we neglect important sources of variability such as spatial heterogeneity in scattering properties and inter-subject differences in skin structure. Beyond the simplicity of the skin model, another limitation of our simulator is its focus on single heartbeats, ignoring both slower dynamics (e.g., respiration) and artifacts such as motion. In addition to forward-model misspecification, PPGen’s capabilities are constrained by the informed prior distribution we introduced. By design, we limited its informativeness to bounding static parameters and loosely structuring what realistic blood waveforms may look like. These assumptions, however, may be overly conservative, since we expect at least some parameters to be correlated in real-world measurements. While our experiments investigate the potential impact of some of these simplifying assumptions, future work should validate our findings on real-world data with clearly defined modeling objectives. Developing rigorous validation strategies for such objectives will be essential to further demonstrate the practical value of PPGen and HAI. 

Taken together, our results show that HAI offers potential for extracting useful health information from real-wold PPG signals, while enabling a degree of clinical interpretability not readily available from either raw signals or traditional black-box ML approaches.
While our results are a proof of concept based on in-silico experiments, future validation on real-world data could establish the unique potential of hybrid-ML tools for leveraging wearable data in health applications.

\bibliography{biblio}

\begin{thebibliography}{50}
\providecommand{\natexlab}[1]{#1}
\providecommand{\url}[1]{\texttt{#1}}
\expandafter\ifx\csname urlstyle\endcsname\relax
  \providecommand{\doi}[1]{doi: #1}\else
  \providecommand{\doi}{doi: \begingroup \urlstyle{rm}\Url}\fi

\bibitem[Abbaspourazad et~al.(2024)Abbaspourazad, Elachqar, Miller, Emrani, Nallasamy, and Shapiro]{abbaspourazad2024largescale}
Salar Abbaspourazad, Oussama Elachqar, Andrew Miller, Saba Emrani, Udhyakumar Nallasamy, and Ian Shapiro.
\newblock Large-scale training of foundation models for wearable biosignals.
\newblock In \emph{The Twelfth International Conference on Learning Representations}, 2024.

\bibitem[Al-Halawani et~al.(2024)Al-Halawani, Qassem, and Kyriacou]{Al-Halawani2024}
Raghda Al-Halawani, Meha Qassem, and Panicos~A Kyriacou.
\newblock Monte carlo simulation of the effect of melanin concentration on light-tissue interactions in transmittance and reflectance finger photoplethysmography.
\newblock \emph{Scientific Reports}, 14\penalty0 (1), 2024.

\bibitem[Alerstam et~al.(2008{\natexlab{a}})Alerstam, Andersson-Engels, and Svensson]{Alerstam2008b}
Erik Alerstam, Stefan Andersson-Engels, and Tomas Svensson.
\newblock White monte carlo for time-resolved photon migration.
\newblock \emph{Journal of biomedical optics}, 13\penalty0 (4), 2008{\natexlab{a}}.

\bibitem[Alerstam et~al.(2008{\natexlab{b}})Alerstam, Svensson, and Andersson-Engels]{Alerstam2008}
Erik Alerstam, Tomas Svensson, and Stefan Andersson-Engels.
\newblock Parallel computing with graphics processing units for high-speed monte carlo simulation of photon migration.
\newblock \emph{Journal of biomedical optics}, 13\penalty0 (6), 2008{\natexlab{b}}.

\bibitem[Benemerito et~al.(2024)Benemerito, Melis, Wehenkel, and Marzo]{Benemerito_2024}
I~Benemerito, A~Melis, Antoine Wehenkel, and A~Marzo.
\newblock openbf: an open-source finite volume 1d blood flow solver.
\newblock \emph{Physiological Measurement}, 45\penalty0 (12), 2024.

\bibitem[Bissiri et~al.(2016)Bissiri, Holmes, and Walker]{bissiri2016general}
Pier~Giovanni Bissiri, Chris~C Holmes, and Stephen~G Walker.
\newblock A general framework for updating belief distributions.
\newblock \emph{Journal of the Royal Statistical Society Series B: Statistical Methodology}, 78\penalty0 (5), 2016.

\bibitem[Bolin et~al.(1989)Bolin, Preuss, Taylor, and Ference]{Bolin1989}
Frank~P Bolin, Luther~E Preuss, Roy~C Taylor, and Robert~J Ference.
\newblock Refractive index of some mammalian tissues using a fiber optic cladding method.
\newblock \emph{Applied optics}, 28\penalty0 (12), 1989.

\bibitem[Boonya-Ananta et~al.(2020)Boonya-Ananta, Rodriguez, Hansen, Hutcheson, and Ramella-Roman]{boonya2020modeling}
Tananant Boonya-Ananta, Andres~J Rodriguez, Anders~K Hansen, Joshua~D Hutcheson, and Jessica~C Ramella-Roman.
\newblock Modeling of a photoplethysmographic (ppg) waveform through monte carlo as a method of deriving blood pressure in individuals with obesity.
\newblock In \emph{Optical Interactions with Tissue and Cells XXXI}, volume 11238. SPIE, 2020.

\bibitem[Cannon et~al.(2022)Cannon, Ward, and Schmon]{cannon2022investigating}
Patrick Cannon, Daniel Ward, and Sebastian~M Schmon.
\newblock Investigating the impact of model misspecification in neural simulation-based inference.
\newblock \emph{arXiv preprint arXiv:2209.01845}, 2022.

\bibitem[Charlton et~al.(2019)Charlton, Mariscal~Harana, Vennin, Li, Chowienczyk, and Alastruey]{Charlton2019}
Peter~H Charlton, Jorge Mariscal~Harana, Samuel Vennin, Ye~Li, Phil Chowienczyk, and Jordi Alastruey.
\newblock Modeling arterial pulse waves in healthy aging: a database for in silico evaluation of hemodynamics and pulse wave indexes.
\newblock \emph{American Journal of Physiology-Heart and Circulatory Physiology}, 317\penalty0 (5), 2019.

\bibitem[Charlton et~al.(2023)Charlton, Allen, Bail{\'o}n, Baker, Behar, Chen, Clifford, Clifton, Davies, Ding, et~al.]{Charlton_2023}
Peter~H Charlton, John Allen, Raquel Bail{\'o}n, Stephanie Baker, Joachim~A Behar, Fei Chen, Gari~D Clifford, David~A Clifton, Harry~J Davies, Cheng Ding, et~al.
\newblock The 2023 wearable photoplethysmography roadmap.
\newblock \emph{Physiological measurement}, 44\penalty0 (11), 2023.

\bibitem[Chatterjee et~al.(2020)Chatterjee, Budidha, and Kyriacou]{Chatterjee2020investigating}
S.~Chatterjee, K.~Budidha, and P.~A. Kyriacou.
\newblock Investigating the origin of {photoplethysmography} using a multiwavelength {Monte Carlo} model.
\newblock \emph{Physiological Measurement}, 41\penalty0 (8), 2020.

\bibitem[Ch{\'e}rief-Abdellatif and Alquier(2020)]{cherief2020mmd}
Badr-Eddine Ch{\'e}rief-Abdellatif and Pierre Alquier.
\newblock Mmd-bayes: Robust bayesian estimation via maximum mean discrepancy.
\newblock In \emph{Symposium on Advances in Approximate Bayesian Inference}. PMLR, 2020.

\bibitem[Cranmer et~al.(2020)Cranmer, Brehmer, and Louppe]{cranmer2020frontier}
Kyle Cranmer, Johann Brehmer, and Gilles Louppe.
\newblock The frontier of simulation-based inference.
\newblock \emph{Proceedings of the National Academy of Sciences}, 117\penalty0 (48), 2020.

\bibitem[Das et~al.(2021)Das, Yuasa, Maeda, Nishidate, Funamizu, and Aizu]{Das2021databaseinversion}
Kaustav Das, Tomonori Yuasa, Takaaki Maeda, Izumi Nishidate, Hideki Funamizu, and Yoshihisa Aizu.
\newblock Simple detection of absorption change in skin tissue using simulated spectral reflectance database.
\newblock \emph{Measurement}, 182, 2021.

\bibitem[Dellaporta et~al.(2022)Dellaporta, Knoblauch, Damoulas, and Briol]{dellaporta2022robust}
Charita Dellaporta, Jeremias Knoblauch, Theodoros Damoulas, and Fran{\c{c}}ois-Xavier Briol.
\newblock Robust bayesian inference for simulator-based models via the mmd posterior bootstrap.
\newblock In \emph{International Conference on Artificial Intelligence and Statistics}. PMLR, 2022.

\bibitem[Doostdar and Khalilzadeh(2014)]{Doostdar2014quantification}
H~Doostdar and MA~Khalilzadeh.
\newblock Quantification the effect of ageing on characteristics of the photoplethysmogram using an optimized windkessel model.
\newblock \emph{Journal of Biomedical Physics \& Engineering}, 4\penalty0 (3), 2014.

\bibitem[Formaggia et~al.(2010)Formaggia, Quarteroni, and Veneziani]{formaggia2010cardiovascular}
Luca Formaggia, Alfio Quarteroni, and Allesandro Veneziani.
\newblock \emph{Cardiovascular Mathematics: Modeling and simulation of the circulatory system}, volume~1.
\newblock Springer Science \& Business Media, 2010.

\bibitem[Frasch et~al.(1996)Frasch, Kresh, and Noordergraaf]{Frasch1996}
H~Frederick Frasch, J~Yasha Kresh, and Abraham Noordergraaf.
\newblock Two-port analysis of microcirculation: an extension of windkessel.
\newblock \emph{American Journal of Physiology-Heart and Circulatory Physiology}, 270\penalty0 (1), 1996.

\bibitem[Fredriksson et~al.(2012)Fredriksson, Larsson, and Str{\"o}mberg]{Fredriksson2012inverse}
Ingemar Fredriksson, Marcus Larsson, and Tomas Str{\"o}mberg.
\newblock Inverse monte carlo method in a multilayered tissue model for diffuse reflectance spectroscopy.
\newblock \emph{Journal of biomedical optics}, 17\penalty0 (4), 2012.

\bibitem[Hale and Querry(1973)]{Hale1973}
George~M Hale and Marvin~R Querry.
\newblock Optical constants of water in the 200-nm to 200-$\mu$ m wavelength region.
\newblock \emph{Applied optics}, 12\penalty0 (3), 1973.

\bibitem[Haque et~al.(2022)Haque, Kwon, and Kim]{Haque2022}
Chowdhury~Azimul Haque, Tae-Ho Kwon, and Ki-Doo Kim.
\newblock Cuffless blood pressure estimation based on monte carlo simulation using photoplethysmography signals.
\newblock \emph{Sensors}, 22\penalty0 (3), 2022.

\bibitem[Huang et~al.(2023)Huang, Bharti, Souza, Acerbi, and Kaski]{huang2023learning}
Daolang Huang, Ayush Bharti, Amauri Souza, Luigi Acerbi, and Samuel Kaski.
\newblock Learning robust statistics for simulation-based inference under model misspecification.
\newblock \emph{Advances in Neural Information Processing Systems}, 36, 2023.

\bibitem[Jacques(2013)]{Jacques2013review}
Steven~L Jacques.
\newblock Optical properties of biological tissues: a review.
\newblock \emph{Physics in Medicine \& Biology}, 58\penalty0 (11), 2013.

\bibitem[Jacques and Wang(1995)]{jacques1995monte}
Steven~L Jacques and Lihong Wang.
\newblock Monte carlo modeling of light transport in tissues.
\newblock In \emph{Optical-thermal response of laser-irradiated tissue}. Springer, 1995.

\bibitem[Jewson et~al.(2018)Jewson, Smith, and Holmes]{jewson2018principled}
Jack Jewson, Jim~Q Smith, and Chris Holmes.
\newblock Principled bayesian minimum divergence inference.
\newblock \emph{Entropy}, 20\penalty0 (6), 2018.

\bibitem[Kelly et~al.(2023)Kelly, Nott, Frazier, Warne, and Drovandi]{kelly2023misspecification}
Ryan~P Kelly, David~J Nott, David~T Frazier, David~J Warne, and Chris Drovandi.
\newblock Misspecification-robust sequential neural likelihood for simulation-based inference.
\newblock \emph{arXiv preprint arXiv:2301.13368}, 2023.

\bibitem[Kyriacou and Allen(2021)]{kyriacou2021photoplethysmography}
P.A. Kyriacou and J.~Allen.
\newblock \emph{Photoplethysmography: Technology, Signal Analysis and Applications}.
\newblock Academic Press, 2021.

\bibitem[Loshchilov and Hutter(2016)]{loshchilov2016sgdr}
Ilya Loshchilov and Frank Hutter.
\newblock Sgdr: Stochastic gradient descent with warm restarts.
\newblock \emph{arXiv preprint arXiv:1608.03983}, 2016.

\bibitem[Loshchilov and Hutter(2017)]{loshchilov2017decoupled}
Ilya Loshchilov and Frank Hutter.
\newblock Decoupled weight decay regularization.
\newblock \emph{arXiv preprint arXiv:1711.05101}, 2017.

\bibitem[Lueckmann et~al.(2017)Lueckmann, Goncalves, Bassetto, {\"O}cal, Nonnenmacher, and Macke]{lueckmann2017flexible}
Jan-Matthis Lueckmann, Pedro~J Goncalves, Giacomo Bassetto, Kaan {\"O}cal, Marcel Nonnenmacher, and Jakob~H Macke.
\newblock Flexible statistical inference for mechanistic models of neural dynamics.
\newblock \emph{Advances in neural information processing systems}, 30, 2017.

\bibitem[Manduchi et~al.(2024)Manduchi, Wehenkel, Behrmann, Pegolotti, Miller, Sener, Cuturi, Sapiro, and Jacobsen]{manduchi2024leveraging}
Laura Manduchi, Antoine Wehenkel, Jens Behrmann, Luca Pegolotti, Andy~C Miller, Ozan Sener, Marco Cuturi, Guillermo Sapiro, and J{\"o}rn-Henrik Jacobsen.
\newblock Leveraging cardiovascular simulations for in-vivo prediction of cardiac biomarkers.
\newblock \emph{arXiv preprint arXiv:2412.17542}, 2024.

\bibitem[Matsubara et~al.(2022)Matsubara, Knoblauch, Briol, and Oates]{matsubara2022robust}
Takuo Matsubara, Jeremias Knoblauch, Fran{\c{c}}ois-Xavier Briol, and Chris~J Oates.
\newblock Robust generalised bayesian inference for intractable likelihoods.
\newblock \emph{Journal of the Royal Statistical Society Series B: Statistical Methodology}, 84\penalty0 (3), 2022.

\bibitem[Meglinski and Matcher(2003)]{Meglinski2003}
Igor~V Meglinski and SJ~Matcher.
\newblock Computer simulation of the skin reflectance spectra.
\newblock \emph{Computer methods and programs in biomedicine}, 70\penalty0 (2), 2003.

\bibitem[Melis(2018)]{melis2018openbf}
A~Melis.
\newblock openbf: Julia software for 1d blood flow modelling.
\newblock \emph{University of Sheffield, Software}, 2018.

\bibitem[Prahl et~al.(1999)]{prahl1999optical}
Scott Prahl et~al.
\newblock Optical absorption of hemoglobin.
\newblock 1999.

\bibitem[Rajaram et~al.(2010)Rajaram, Gopal, Zhang, and Tunnell]{Rajaram2010}
Narasimhan Rajaram, Ashwini Gopal, Xiaojing Zhang, and James~W Tunnell.
\newblock Experimental validation of the effects of microvasculature pigment packaging on in vivo diffuse reflectance spectroscopy.
\newblock \emph{Lasers in surgery and medicine}, 42\penalty0 (7), 2010.

\bibitem[Reiser et~al.(2022)Reiser, Breidenassel, and Amft]{Reiser2022}
Maximilian Reiser, Andreas Breidenassel, and Oliver Amft.
\newblock Simulation framework for reflective ppg signal analysis depending on sensor placement and wavelength.
\newblock In \emph{2022 IEEE-EMBS International Conference on Wearable and Implantable Body Sensor Networks (BSN)}. IEEE, 2022.

\bibitem[Scarbrough et~al.(2024)Scarbrough, Chen, and Yu]{Scarbrough2024mlInversion}
Allison Scarbrough, Keke Chen, and Bing Yu.
\newblock Designing a use-error robust machine learning model for quantitative analysis of diffuse reflectance spectra.
\newblock \emph{Journal of Biomedical Optics}, 29\penalty0 (1), 2024.

\bibitem[Sekar et~al.(2017)Sekar, Bargigia, Mora, Taroni, Ruggeri, Tosi, Pifferi, and Farina]{Sekar2017}
Sanathana Konugolu~Venkata Sekar, Ilaria Bargigia, Alberto~Dalla Mora, Paola Taroni, Alessandro Ruggeri, Alberto Tosi, Antonio Pifferi, and Andrea Farina.
\newblock Diffuse optical characterization of collagen absorption from 500 to 1700 nm.
\newblock \emph{Journal of biomedical optics}, 22\penalty0 (1), 2017.

\bibitem[Takeishi and Kalousis(2021)]{takeishi2021physics}
Naoya Takeishi and Alexandros Kalousis.
\newblock Physics-integrated variational autoencoders for robust and interpretable generative modeling.
\newblock \emph{Advances in Neural Information Processing Systems}, 34, 2021.

\bibitem[Tamura(2019)]{tamura2019current}
Toshiyo Tamura.
\newblock Current progress of photoplethysmography and spo2 for health monitoring.
\newblock \emph{Biomedical engineering letters}, 9\penalty0 (1):\penalty0 21--36, 2019.

\bibitem[Tanaka(2022)]{Tanaka_2022}
Akio Tanaka.
\newblock Analysis of a microcirculatory windkessel model using photoplethysmography with green light: A pilot study.
\newblock \emph{IEICE Electronics Express}, 19\penalty0 (21), 2022.

\bibitem[Tang et~al.(2020)Tang, Chen, Ward, and Elgendi]{Tang2020synthetic}
Qunfeng Tang, Zhencheng Chen, Rabab Ward, and Mohamed Elgendi.
\newblock Synthetic photoplethysmogram generation using two gaussian functions.
\newblock \emph{Scientific Reports}, 10\penalty0 (1), 2020.

\bibitem[Van~Veen et~al.(2000)Van~Veen, Sterenborg, Pifferi, Torricelli, and Cubeddu]{vanVeen2004}
RLP Van~Veen, HJCM Sterenborg, A~Pifferi, A~Torricelli, and R~Cubeddu.
\newblock Determination of vis-nir absorption coefficients of mammalian fat, with time-and spatially resolved diffuse reflectance and transmission spectroscopy.
\newblock In \emph{Proceedings of Biomedical Topical Meetings}, 2000.

\bibitem[Van~Veen et~al.(2002)Van~Veen, Verkruysse, and Sterenborg]{vanVeen2002}
RLP Van~Veen, W~Verkruysse, and HJCM Sterenborg.
\newblock Diffuse-reflectance spectroscopy from 500 to 1060 nm by correction for inhomogeneously distributed absorbers.
\newblock \emph{Optics letters}, 27\penalty0 (4), 2002.

\bibitem[Wang et~al.(1995)Wang, Jacques, and Zheng]{wang1995mcml}
Lihong Wang, Steven~L Jacques, and Liqiong Zheng.
\newblock Mcml—monte carlo modeling of light transport in multi-layered tissues.
\newblock \emph{Computer methods and programs in biomedicine}, 47\penalty0 (2), 1995.

\bibitem[Ward et~al.(2022)Ward, Cannon, Beaumont, Fasiolo, and Schmon]{ward2022robust}
Daniel Ward, Patrick Cannon, Mark Beaumont, Matteo Fasiolo, and Sebastian Schmon.
\newblock Robust neural posterior estimation and statistical model criticism.
\newblock \emph{Advances in Neural Information Processing Systems}, 35, 2022.

\bibitem[Wehenkel et~al.(2025)Wehenkel, Gamella, Sener, Behrmann, Sapiro, Jacobsen, and marco cuturi]{wehenkel2024addressing}
Antoine Wehenkel, Juan~L. Gamella, Ozan Sener, Jens Behrmann, Guillermo Sapiro, Joern-Henrik Jacobsen, and marco cuturi.
\newblock Addressing misspecification in simulation-based inference through data-driven calibration.
\newblock In \emph{Forty-second International Conference on Machine Learning}, 2025.

\bibitem[Zhong et~al.(2014)Zhong, Wen, and Zhu]{Zhong2014reflectanceInversion}
Xiewei Zhong, Xiang Wen, and Dan Zhu.
\newblock Lookup-table-based inverse model for human skin reflectance spectroscopy: two-layered monte carlo simulations and experiments.
\newblock \emph{Optics express}, 22\penalty0 (2), 2014.

\end{thebibliography}

\newpage
\clearpage
\onecolumn
\appendix

\section{Notations}

\begin{table*}[h]
\centering
\caption{Summary of notation. }\label{tab:notations}
\renewcommand{\arraystretch}{1.2}
\begin{tabular}{ll}
\toprule
Symbol & Description \\
\midrule
$\mathbf{x}_o \in \mathbb{R}^{R \times N \times T}$ & Observed PPG pulse with $R$ receivers, $N$ emitters, $T$ timesteps \\

$\mathbf{\hat{x}}_s \in \mathbb{R}^{R \times N \times T}$ & Noiseless simulated (synthetic) PPG pulse generated by the forward model \\
$\mathbf{x}_s \in \mathbb{R}^{R \times N \times T}$ & Simulated (synthetic) PPG pulse generated by the forward model \\
$\mathbf{x}^t_s \in \mathbb{R}^{R \times N}$ & Simulated PPG at timestep $t$ \\
\addlinespace
$\theta \in \Theta \subset \mathbb{R}^{9+2T}$ & Biophysical parameters (static + dynamic) \\
$\theta_s \in \mathbb{R}^9$ & Static parameters (e.g., melanin, blood fractions, vessel diameters, oxygenation) \\
$\theta_d = (\theta^1_d, \ldots, \theta^T_d),\;\theta^t_d \in \mathbb{R}^2$ & Dynamic parameters (blood volume changes per timestep) \\
$\theta^t = [\theta_s, \theta^t_d] \in \mathbb{R}^{11}$ & Biophysical parameters at timestep $t$ \\
$\lambda \in \mathbb{R}$ & Wavelength of emitted light \\
\addlinespace
$\mu_a(\lambda) \in \mathbb{R}^l$ & Absorption coefficients at wavelength $\lambda$ in $l$ skin layers \\
$\mu_s(\lambda) \in \mathbb{R}^l$ & Scattering coefficients at wavelength $\lambda$ in $l$ skin layers \\
$f_b(\theta^t, \lambda)$ & Mapping from biophysical parameters to optical properties $(\mu_a,\mu_s)$ \\
$\hat f_{\text{LT}}(\mu_a, \mu_s)$ & Surrogate light transport simulator: maps optical properties to noiseless PPG \\
$f(\theta)$ & Full differentiable forward model $f = \hat f_{\text{LT}} \circ f_b$ \\
\addlinespace
$p(\mathbf{x}_s \mid \theta)$ & Likelihood of simulated PPG given parameters \\
$p(\mathbf{x}_o \mid \theta)$ & Likelihood of observed PPG given parameters \\
$p(\theta)$ or $\pi(\theta)$ & Prior over biophysical parameters \\
$p(\theta, \mathbf{x}_s) = \,p(\mathbf{x}_s \mid \theta) \pi(\theta)$ & Joint generative model \\
$q_\psi(\mathbf{x}_s \mid \mathbf{x}_o)$ & Misspecification (correction) model, parameterized by $\psi$ \\
$q_\phi(\theta \mid \mathbf{x}_s)$ & Neural posterior estimator of $\theta$ given synthetic $\mathbf{x}_s$ \\
$q^\star_\phi(\theta \mid \mathbf{x}_s)$ & Optimal neural posterior estimator (limit of training) \\
$p(\theta \mid \mathbf{x}_o)$ & True posterior distribution of parameters given observed data \\
\addlinespace
$k_s, \sigma_w^2$ & Shot noise and white noise parameters of sensor model \\
$\mathcal{N}(\mu,\sigma^2)$ & Normal distribution with mean $\mu$ and variance $\sigma^2$ \\
\addlinespace
$\mathbb{E}_{q(\cdot)}[\cdot]$ & Expectation under distribution $q$ \\
$\mathcal{L}$ & Loss function \\
\bottomrule
\end{tabular}
\end{table*}

\clearpage

\section{PPGen: PPG Pulse Generator}\label{app:PPGen}
\subsection{Prior over biophysical parameters}
\label{app:prior}

A detailed list of the biophysical parameters $\theta = \left[ \theta_s, \theta_d \right]$ considered in our model can be found in \autoref{tab:PSRs}.
\begin{table*}[h]
    \centering
    \begin{tabular}{ccp{.55\textwidth}ccc}
       \toprule
          & \bfseries Name & \bfseries Description & \bfseries Range & \bfseries Unit & \bfseries Type\\
        \midrule
        \multirow{2}{*}{$\mu_s$} 
        & A & Scattering amplitude parameter. & [0.25 - 1.0] & mm$^{-1}$ & Static\\
        & SP & Scattering power parameter. & [1.3 - 1.5] & – & Static\\ \midrule 
        
        \multirow{9}{*}{\vspace{-8em}$\mu_a$} &
       $\text{Mel}$ & Melanin fraction in the epidermis. & [0.25 - 14] & \% & Static\\
        & $\text{BV}_2$ & Blood fraction during diastole in the dermis. It informs about the tissue perfusion. & [0.1 - 4] & \% & Static\\
        & $\text{BV}_3$ & Blood fraction during diastole in the subcutaneous layer. It informs about the tissue perfusion. & [0.1 - 8] & \% & Static\\
        & $\text{VD}_2$ & Vessel diameter in the dermis. It is an aging biomarker. &  [0.01 - 0.04] & $\text{mm}$ & Static\\
        
        & $\text{VD}_3$ & Vessel diameter in the subcutaneous layer. It is an aging biomarker. &  [0.04 - 0.06]&  $\text{mm}$ & Static\\
        
        & SA & Arterial oxygen saturation. It informs on the oxygenation status (e.g., hypoxemia). & [60 - 100] & \% & Static\\
        & $\Delta\text{SV}$ & Arteriovenous oxygen saturation difference. It informs on the oxygen extraction (e.g., exercise, sepsis). & [1 - 20]& \% & Static\\
        
        & $\Delta\text{BV}_2$ & Systolic blood fraction scaling in dermis. It informs on the cardiovascular system behavior. & [1.0 - 1.02] & – & Dynamic\\
        & $\Delta\text{BV}_3$ & Systolic blood fraction scaling in subcutaneous layer. It informs on the cardiovascular system behavior. & [1.0 - 1.02] & – & Dynamic\\
        \bottomrule
    \end{tabular}
    \caption{List of biophysical parameters considered in our model. While most parameters can be considered static within short time windows ($\approx 1$ second), blood volume related changes occur in much shorter time-scales and are considered dynamic in our pipeline. 
    }
    \label{tab:PSRs}
\end{table*}

The set of biophysical parameters is sampled from an informed prior $\pi(\theta)$. We describe below the two different strategies used to obtain samples for static parameters $\theta_s$ and for dynamic parameters $\theta_d$.

\subsubsection{Static parameter sampling}
\label{app:static_psr}

Static biophysical parameters $\theta_s$ are sampled uniformly from a range of values considered biophysically plausible and extracted from the literature. The ranges considered for each parameter are provided in \autoref{tab:PSRs}.

\subsubsection{Blood volume waveform generation}
\label{app:blood_volume_sim}
Our priors for dynamic tissue properties $\theta_d$, namely, the blood volume waveforms in the dermis and subcutaneous tissue layers, are based on mechanistic models. The dynamic component of tissue blood volume is a result of pulsatile pressure waves traveling from large underlying arteries through a network of smaller vessels (capillaries, arterioles and small arteries) in the tissue before draining into the venous system. For our purposes, we use synthetic arterial pressure waves simulated at the radial artery of the (left) wrist, and map them to blood volume waveforms proximal to the skin at the same location.

\paragraph{Synthetic arterial pressure waveforms.} We consider the dataset of arterial pressure waves introduced in \cite{manduchi2024leveraging}. These waveforms were simulated using the finite-volume solver described in \cite{melis2018openbf, Benemerito_2024} by randomly sampling parameters that influence the cardiovascular system function of virtual subjects (heart rate, stroke volume, peak flow time, reverse flow volume, subject height) in physiological ranges. See \cite{Charlton2019} for details on how these virtual subjects were defined. Although the solver allows for the evaluation of pressure waveforms at any point in the human body, we chose to only consider those simulated at a random location of the left radial artery (the vessel supplying blood to the left wrist) because of its relevance in wearable devices such as smartwatches. More specifically, the dataset consists of $N_s=81'660$ waveforms of the form $[P_{1,j},\ldots,P_{T,j}]$, for $j = 1,\ldots,N_s$, where $P_{i,j}$ is the reading of the arterial pressure wave of the virtual subject $j$ at the sample point $i$. As all waveforms are resampled over a common grid composed of $T = 64$ points, we also keep track of the sampling frequency $f_j = T / M_{j}$, where $M_{j}$ is the number of sample points per cardiac cycle in the original waveform (note that, in the original dataset, all waveforms are stored with a common sampling frequency of 500 Hz).

\begin{figure*}
    \centering
    \includegraphics[width=1\linewidth]{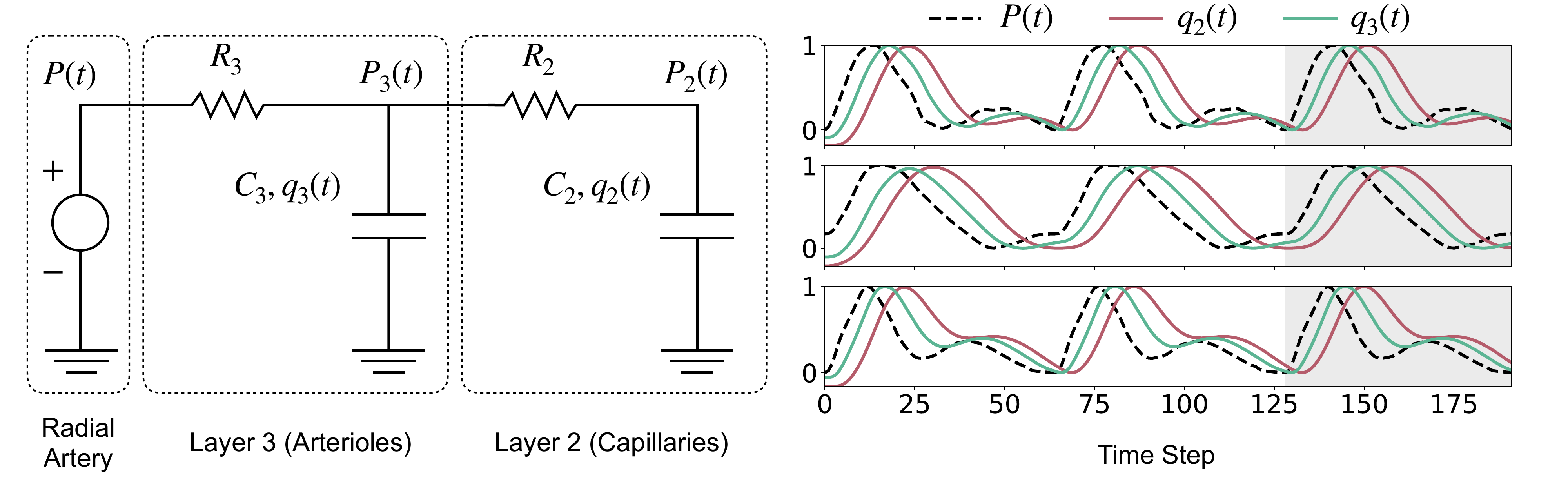}
    \caption{On the left, Windkessel model relating the arterial pressure wave at the radial artery $P(t)$ to blood volumes in the two layers of the skin $q_2(t)$ and $q_3(t)$. On the right, solution of the Windkessel equations for three different virtual subjects over three cardiac cycles. The shaded area represents the last cardiac cycle, which would be used to extract the blood volumes to use in our model.}
    \label{fig:windkessel}
\end{figure*}

\paragraph{Blood volume waveforms.} Given the arterial pressure waveforms described above, 
we use Windkessel models to compute blood volume waveforms proximal to the skin. For a reference of the application of Windkessel models in the microvasculature see e.g. \citet{Frasch1996}. These models simplify the fluid dynamics of blood flow by reducing a cardiovascular system of vessels to an analogous lumped parameter electrical circuit.

In our case, we consider the two-layer model shown in \autoref{fig:windkessel} (left), with three compartments that represent the radial artery, narrower vessels in subcutaneous (Layer 3 in our tissue model), and even narrower vessels in dermis (Layer 2). Pressure in the radial artery is equivalent to the potential $P(t)$ at the voltage source on the left. The two layers act as a low-pass RC filters with time constants $\tau_3 = R_3C_3$ and $\tau_2 = R_2C_2$ respectively. Note that Layer 1 (not shown in \autoref{fig:windkessel}) representing the top epidermis layer is not included in the model due to absence of any significant vessels in its structure. The blood volumes in layers 3 and 2 are equivalent to the charges $q_3(t)$ and $q_2(t)$ on the respective capacitors. Eventually, the mathematical relationships between $P(t)$, $q_3(t)$ and $q_2(t)$ are second-order differential equations of the form
\begin{equation}
    \begin{aligned}
        P(t) &\approx \dfrac{\tau_2 \tau_3}{C_2} \dfrac{\text{d}^2 q_2}{\text{d}t^2} + \dfrac{\tau_2 + \tau_3}{C_2}\dfrac{\text{d}q_2}{\text{d}t}  + \dfrac{1}{C_2}q_2(t),\\
        \tau_2 C_3 \frac{\text{d}P(t)}{\text{d}t} + P(t) &\approx \frac{\tau_3 \tau_2}{C_3} \frac{\text{d}^2q_3(t)}{\text{d}t^2} + \frac{\tau_3 + \tau_2}{C_3} \frac{\text{d}q_3(t)}{\text{d}t}  + \frac{1}{C_3} q_3(t).
    \end{aligned}
    \label{eq:windkessel}
\end{equation}
To create blood volume waveforms, we solve \eqref{eq:windkessel} starting from batches of the arterial pressure waves by sampling a set of indices $ \mathcal{J} \in \{1,\ldots, N_s\}$ with $|\mathcal{J}| = n_\text{bs}$ is the batch size. For each batch, we randomly sample system parameters $R_2$, $C_2$, $R_3$, $C_3$, while enforcing physiological conditions based on realistic assumptions on the radii in the different compartments. We recall that, if $R$ and $C$ are the resistance and capacitance of a blood vessel and $r$ is its radius, then $R \propto r^{-4}$ and $C \propto r^3$ (see \cite{formaggia2010cardiovascular}). Since vessel radii decrease from deeper to shallower tissue, our sampling strategy enforces
\begin{equation}
    \begin{aligned}
        R_2 &> R_3 \\
        C_2 &< C_3 \\
        \tau_2 = R_2 C_2 &> \tau_3 = R_3 C_3,
    \end{aligned}
\end{equation}
where $\tau_2$ and $\tau_3$ are the time constants of the two compartments. In order to solve the continuous-time dynamics we approximate \eqref{eq:windkessel} in discrete time using pole-zero matching, and recursively compute the blood volume waveforms at time step size $\Delta t$. Given the notation introduced in the previous paragraph, the blood volume  waveform $q_{2,i,j}$ at time step $i$ for virtual subject $j \in \mathcal{J}$ can be expressed as
\begin{equation}
    \begin{aligned}
        q_{2,i,j} &= P_{i-2,j} \cdot C_2 \cdot \big(1 - e^{-\Delta t / \tau_2}\big) \cdot \big(1 - e^{-\Delta t / \tau_3}\big) \\
        &\quad + q_{2,i-1,j} \cdot \big(e^{-\Delta t / \tau_2} + e^{-\Delta t / \tau_3}\big) \\
        &\quad - q_{2,i-2,j} \cdot \big(e^{-\Delta t / \tau_2} \cdot e^{-\Delta t / \tau_3}\big),
    \end{aligned}
    \label{eq:q2}
\end{equation}
where $\Delta t = 1/f_j$ is the time step size. Similarly, for Layer 3, the blood volume waveform $q_{3,i,j}$ can be expressed as
\begin{equation}
    \begin{aligned}
        q_{3,i,j} &= \big(P_{i-1} - e^{-\Delta t / \tau_2} \cdot P_{i-2}\big) \cdot C_3 \cdot \big(1 - e^{-\Delta t / \tau_3}\big) \\
        &\quad + q_{3,i-1,j} \cdot \big(e^{-\Delta t / \tau_2} + e^{-\Delta t / \tau_3}\big) \\
        &\quad - q_{3,i-2,j} \cdot \big(e^{-\Delta t / \tau_2} \cdot e^{-\Delta t / \tau_3}\big).
    \end{aligned}
    \label{eq:q3}
\end{equation}
To use these equations in practice, we need to assign initial conditions to $q_{2,i,j}$ and $q_{3,i,j}$. In particular, we choose zero initial condition as
\begin{equation*}
    q_{2,1,j} = q_{2,2,j} = q_{3,1,j} = q_{3,2,j} = 0,
\end{equation*}
for all $j \in {\mathcal{J}}$. Due to this strong approximation, we allow the system to reach convergence (i.e., a quasi-periodic behavior over a single heart beat) by simulating multiple cardiac cycles. This is possible because the arterial pressure waves are designed to be approximately periodic, namely $P_{i,j} \approx P_{T-1,j}$ for $j = 1,\ldots,N_s$, and the same arterial pressure wave can be used for all cardiac cycles. \autoref{fig:windkessel} (right) shows the solution of \autoref{eq:q2} and \autoref{eq:q3} for three virtual subjects over three cardiac cycles.

Once $q_2(t)$ and $q_3(t)$ are computed by the solver, they are used as the shape of the blood volume waveforms $\Delta\text{BV}_2$ and $\Delta\text{BV}_3$. Afterwards, the minimum and maximum for both waveforms are sampled in $[1.0, 1.02]$, with the constrain that $\text{max}(\Delta \text{BV}) - \text{min}(\Delta \text{BV}) \geq 1.01$. This constraint establishes a lower bound on the amplitude of a blood volume pulse. 

\subsection{Optical tissue model}~\label{app:PSR_map}

Once the biophysical parameters $\theta$ have been sampled from the prior $\pi(\theta)$, we use an optical tissue model to compute the corresponding optical properties. The optical properties of the tissue given a set of parameters $\theta$ consist of the layer-specific absorption coefficient $\mu_a$ and scattering coefficient $\mu_s$. Note that, while $\mu_a$ is layer-specific, we simplify our tissue model and assume $\mu_s$ is shared across all tissue layers.
The modeling approaches for absorption and scattering differ significantly and are explained below. 

\subsubsection{Absorption coefficient $\mu_a$}
The absorption coefficient, $\mu_a(\lambda) \in \mathbb{R}$, scales with the molar extinction coefficient, $\varepsilon_k(\lambda) \in \mathbb{R}$, as a weighted sum
\begin{align*}
    \mu_a(\lambda) = \sum_{k=1}^K \varepsilon_k(\lambda) c_k.
\end{align*}
The molar concentrations $c_k \in \mathbb{R}$ of the dominating $K$ light-absorbing molecules (chromophores) in each tissue layer, primarily oxygenated hemoglobin, deoxygenated hemoglobin, and melanin, are drawn from established extinction spectra from the following references: 
\begin{itemize}
    \item Hemoglobin \citep{prahl1999optical}
    \item Water \citep{Hale1973}
    \item Lipid \citep{vanVeen2004}
    \item Collagen \citep{Sekar2017}
    \item Melanin \citep{Jacques2013review}.
\end{itemize}
For a visualization of extinction spectra, see for example Figure 1 in \citet[]{Meglinski2003} .

We consider different compositions across tissue layers. In particular, we consider the following six chromophores: hemoglobin (oxygenated and de-oxygenated), Water, Lipids, Collagen, and Melanin. See \autoref{tab:choromophores} for an overview of concentrations across tissue layers.

For hemoglobin (blood) we consider a more detailed model as it is the key chromophore that lets us understand the cardiovascular system. First, we define arterial and venous blood fraction with a fixed static ratio of 1 / 3 (arterial / venous). Using this and the arterial oxygen saturation (SA) and venous oxygen saturation ($\text{SV} = \text{SA} - \Delta \text{SV}$), we define the concentration of oxygenated and de-oxygenated blood as
\begin{align*}
    c_{oxy} &= \frac{1}{4} \cdot \frac{\text{SA}}{100} \cdot \text{BV} \cdot\Delta \text{BV} +  \frac{3}{4}\cdot \frac{\text{SV}}{100}\cdot \text{BV} \\
    c_{deoxy} &= \frac{1}{4}\left(1 - \frac{\text{SA}}{100}\right) \text{BV}\cdot \Delta \text{BV} + \frac{3}{4} \left(1 - \frac{\text{SV}}{100}\right) \text{BV}, 
\end{align*}
where the first summand in each equation corresponds to the contribution of arterial and the second to venous blood. The component $\Delta \text{BV}$ represents the systolic blood scaling factor (blood volume waveforms). Our simulation model for these parameters is described in \autoref{app:blood_volume_sim}.

Furthermore, we employ a vessel diameter (denoted as VD) based correction to include the effects of mean blood vessel size of the absorption of blood as
\begin{align*}
    \tilde{\mu_a}^{blood} = \frac{\text{BV} \cdot \Delta \text{BV} \cdot (1 - e^{- \frac{\mu_a^{blood}}{\text{BV}} \cdot \Delta \text{BV} \cdot \text{VD}})}{\text{VD}}.
\end{align*}
This correction was proposed by \cite{vanVeen2002} and later validated by \cite{Rajaram2010}.

Lastly, we share the arterial (SA) and venous oxygen saturation (SV) across the dermis and subcutaneous layer, while blood fraction and blood volume waveforms are considered separately for both layers. For an overview of all physiological parameters see \autoref{tab:PSRs}.

\begin{table*}[]
    \centering
    \begin{tabular}{cccc}
        \toprule
        \bfseries Name & \bfseries Epidermis & \bfseries Dermis & \bfseries Subcutaneous \\
        \midrule
       Hemoglobin (oxygenated) & 0 & vary & vary \\
       Hemoglobin (de-oxygenated) & 0 & vary & vary \\
       Water & 60 & 70 & 10 \\
       Lipid & 35 & 5 & 90 \\
       Collagen & 0 & 25 & 0 \\
       Melanin & vary & 0 & 0 \\
       \bottomrule
    \end{tabular}
    \caption{Overview of considered chromophores for the three considered tissue layers, where each value represents a volume fraction in \%. Some concentrations are not fixed to a specific value (denoted by \textit{vary}) and are thus (indirectly) inferred using our parameters $\theta$ as described in \autoref{app:PSR_map}.
    }
    \label{tab:choromophores}
\end{table*}

\subsubsection{Scattering coefficient $\mu_s$}
In contrast to absorption $\mu_a$, the scattering coefficient $\mu_s$ is modeled empirically. While scattering is influenced by biological factors like cell density and size, the relationship is less direct. We therefore adopt a standard empirical formulation that effectively models the observed wavelength-dependence of tissue scattering \citep{Jacques2013review}. More specifically we model the scattering parameter $\mu_s(\lambda)$ as
\begin{align*}
    \mu_s(\lambda) = \frac{\mu_s'}{1 - g},
\end{align*}
where $g \in \mathbb{R}$ is a fixed anisotropy coefficient and $\mu_s' \in \mathbb{R}$ is the reduced scattering coefficient, modeled via a power law as
\begin{align*}
    \mu_s'(\lambda) = \text{A} \left(\frac{\lambda}{1000}\right) ^{-\text{SP}},
\end{align*}
where wavelength $\lambda$ is represented in nm units with $1000$ being the base wavelength, $\text{A}$ describes the scattering amplitude and $\text{SP}$ the scattering power parameter. 

In our tissue model we consider three tissue layers (epidermis, dermis, and subcutaneous layer). While most tissue properties vary from individual to individual, we fixed the following parameters to simplify the model. For the epidermis layer we employ a layer thickness of 0.2 mm, for the dermis 1.5 mm, and for the subcutaneous layer we use 18.3 mm. In terms of scattering, we assume that scattering properties are the same across tissue layers. We use a refractive index of 1.4 for each layer as proposed by \cite{Bolin1989}. Lastly, we employ the Henyey-Greenstein phase function with $g=0.9$ to represent the anisotropy of scattering, which is consistent with reported tissue values \citep{Jacques2013review}.

\subsection{Building a surrogate model of light transport}\label{app:surrogate_training}

\paragraph{Simulation settings.}
We build surrogate models of the computationally expensive light transport simulations with the following key settings for the simulation:
\begin{itemize}
    \item $10^7$ photons for each simulation
    \item Post-processing from scattering-only base simulation as proposed by \cite{Alerstam2008b}
    \item Wavelength-independence assumption with average properties of the sensor probe (reflectivity, refractive index).
\end{itemize}
As stated above, for simplicity we assume that light propagation does not depend on the wavelength and is fully governed by absorption $\mu_a$ and scattering $\mu_s$. This assumption may be violated in practice if the material of the sensor has wavelength-dependent reflection properties, but it allows us to reduce the computational burden significantly. In particular, we can build a surrogate model $\hat{f}_{\text{LT}}: (\mu_a, \mu_s) \mapsto \mathbf{x}_s^t$, which is independent of the wavelength. 

\paragraph{Simulation lookup table.}
This property is then used to create a training dataset for the surrogate model, which we call \textit{simulation lookup table} (LUT). For this dataset we first define the ranges for $\mu_a$ and $\mu_s$ per tissue layer by taking the ranges for the tissue parameters (see \autoref{tab:PSRs}) and using the optical tissue model (see \autoref{app:PSR_map}) to compute maximal and minimal values. In summary, this step yields min/max values in a 4 dimensional space (absorption for three tissue layers and a shared scattering value for all tissue layers). 

To sample this 4D-space, we apply a log-transform to the absorption values, which over-weights lower values during training. This approach is motivated by the extinction properties of the considered chromophores. For instance, hemoglobin's extinction coefficient is very low for wavelengths above 600nm, meaning changes in its concentration produce only minor perturbations in absorption in this range.

We then employ various sampling strategies to form the LUT:
\begin{itemize}
    \item Sobol Sampling (quasi-random sampling to ensure guaranteed coverage)
    \item Two independent Latin Hypercube samples (to balance randomness with coverage of the space).
\end{itemize}
Furthermore, we aim to increase sensitivity in our surrogate model towards small blood-induced perturbations as this is the key chromophore we want to measure with PPG sensors. For this, we add 5 different sets of perturbations in absorption per layer, where we use random multiplier of $\mu_a$ with standard deviations from $[ 0.00001, 0.0001, .001, 0.01, 0.1, 0.0]$.

In summary, the LUT consists of \begin{itemize}
    \item 35 different values for $\mu_s$
    \item 10k samples for each Sobol and Latin Hypercube sample
    \item 6 $\mu_a$ perturbations for each base simulation.
\end{itemize}
This yields 5.25 million datapoints that can be used to train a surrogate model of the computationally expensive light simulation. 

\paragraph{Training settings for the surrogate model.}
After constructing a simulation lookup table (LUT), we define the surrogate model of light transport as a mapping
\begin{align*}
    \hat{f}_{LT}: (\mu_a^1, \mu_a^2, \mu_a^3, \mu_s) \mapsto \mathbf{x}_s^t,
\end{align*}
where $\mu_a^l$ is the absorption in tissue layer $l$ and $\mathbf{x}_s^t \in \mathbb{R}^4$ as we consider 4 distinct receiver channel (paths between single light emitter and 4 detectors). As the mapping may be highly non-linear and due to having access to a large LUT, we employ neural networks for $\hat{f}_{LT}$ with the following workflow: 
\begin{enumerate}
    \item Log-transform of $\mu_a$ and $\mathbf{x}_s^t$
    \item Standardization of inputs and outputs based on mean/std computed over 10 batches from the training set
    \item Multi-layer perceptron with 3 layers of size $[100, 100, 100]$ and $\tanh$  nonlinearity to guarantee smoothness of the predictions.
\end{enumerate}
After defining the architecture, we split the LUT intro training and validation dataset by leaving 15\% of $\mu_s$ values for validation. We split across $\mu_s$, because we have fewer diversity in scattering due to the computationally expensive simulation w.r.t. scattering. We use evenly spaced intervals. This ensures the validation groups are spread out across the entire range of $\mu_s$ values, which is useful as $\mu_s$ represents a continuous physical parameter value and we want to ensure the the validation set tests the model's performance across that entire range.

As training losses, we use mean-absolute-error (MAE) to reduce the effect of outliers from the LUT (which may happen due to the stochastic nature of the Monte-Carlo light transport simulation). We train the surrogate model $\hat{f}_{LT}$ using both a MAE loss of amplitudes as well as a MAE loss on amplitude differences for $\mu_a$ base simulations and perturbation cases (see description of dataset in previous paragraph). Both losses are weighted equally after normalizing each loss component with the loss value at from the first batch.
    
We then train with a batch size of 1000, a learning rate of 1e-4, Adam optimizer for 400 epochs. The model is then selected based on the lowest loss on the validation set.

\subsection{Sensor model and noise model}
\label{app:sensor_and_noise_model}

\paragraph{Sensor Model.}
As described in \autoref{sec:noise}, we consider a common four-wavelength PPG setup and a wide-spectrum spectroscopic device as two test cases with the following characteristics (see \autoref{fig:overview_pipeline} for a visual sketch of the sensor): 
 \begin{itemize}
     \item Four-wavelength PPG setup with $N=4$ wavelengths (green, red and two infrared LEDs) and $R=4$ channels to simulate different spacings between LEDs and photodetectors.
     \item Wide-spectrum spectroscopic sensor with $N=531$ (470nm - 1000nm) and $R=4$ as above. 
\end{itemize}
To make the comparison fair, we consider the same geometrical layout of light source and detector for both cases. In particular, we have $R=4$ channels that are obtained from a single light source and four distinct spacings of detectors. The spacings are 3/4/5/6 mm, the center illumination source is 3.2 mm and detection fibers are approximately 0.5mm. 

To model the four-wavelength PPG sensor, we employ profiles over the wavelength range to mimic the emission of lights from LEDs using in PPG sensors. Since LEDs are rather emitting light over a range of wavelengths, each light source emits a distribution of photons with different wavelengths. Formally, for a PPG signal at timestep $t$ from channel $r$ and color $n$, we apply a weighted sum
\begin{align*}
    \mathbf{x}_s^{r,n,t} = \sum_{i} \mathbf{x}_s^{r,t}(\lambda_i) L_n(\lambda_i),
\end{align*}
with LED profiles $\sum_{i} L_n(\lambda_i) = 1$. In our four-wavelength sensor setup, we considered LED profiles centered at 525 nm (green), 660nm (red), 850 nm (infrared), and 940nm (infrared). 

\paragraph{Noise Model.}
We apply noise independently to each scalar $\hat{\mathbf{x}}_s^{r,n,t}$ in the signal tensor $\hat{\mathbf{x}}_s \in \mathbb{R}^{R \times N \times T}$. First, we represent shot noise as zero-mean Gaussian noise with signal-dependent variance $\eta_{Shot} \sim \mathcal{N}(0,\, k_s \cdot \hat{\mathbf{x}}_s^{r,n,t})$, where $k_s$ controls the magnitude of shot noise. Second, we model additive Gaussian noise $\eta_w \sim \mathcal{N}(0,\, \sigma_w^2)$, representing sensor-internal effects such as thermal and chip noise. After sampling these two noise components, we add both components to the clean signal. To cover multiple levels of noise, we explore various values for $k_{shot}$ and $\sigma_w$, see \autoref{tab:noise_levels}.

\begin{figure}
    \centering
    \includegraphics[width=1.\linewidth]{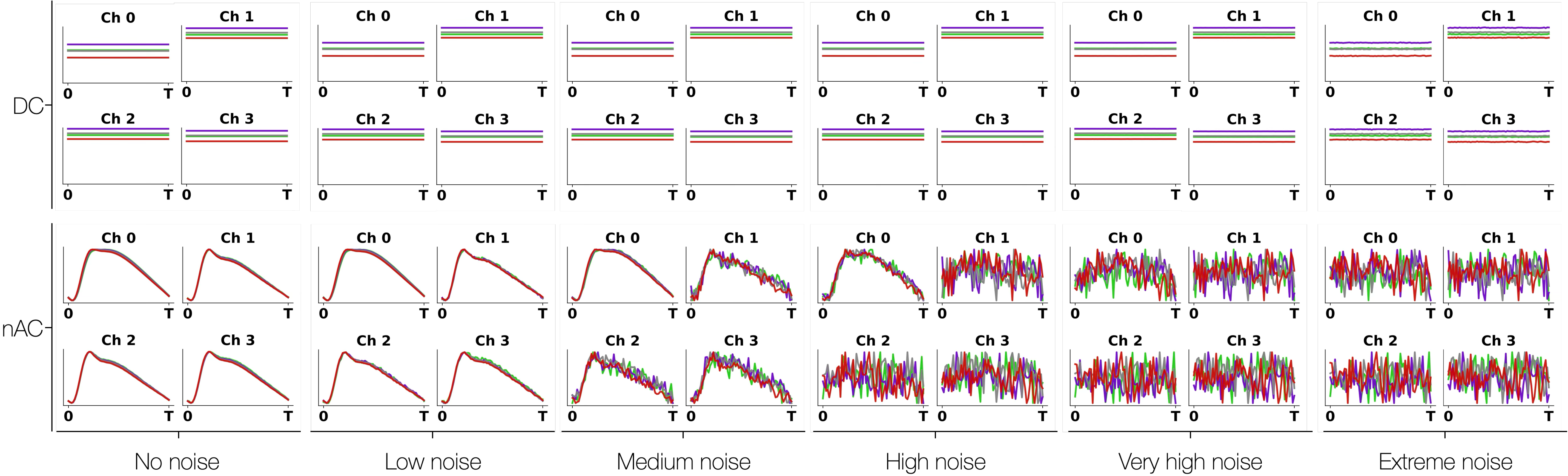}
    \caption{Vizualization of the six noise levels considered in our experiments for a randomly selected four-wavelength measurement. The effect of the noise is mostly noticeable on the normalized AC component of PPG measurements. Nevertheless, it challenges parameter inference as shown in \autoref{fig:clean_results}.}
    \label{fig:noise_viz}
    \vspace{-1em}
\end{figure}

\paragraph{Artificial misspecification.} We construct multiple misspecification settings as discussed in \autoref{results:misspec}. The \textbf{Noise} misspecification uses clean signals for the simulated data $\mathbf{x}_s$ and medium noise for the observed data $\mathbf{x}_o$. For \textbf{Sensor} misspecification we add an offset of 0.1 to each receiver channel, before normalization (see details on normalization in \autoref{app:exp_details}). The \textbf{Skin} misspecification is obtained by training another surrogate model $\tilde{f}_{LT}$ using a simulation lookup table that was generated with a slightly perturbed layer thickness. The layer thickness for this perturbed setting was 0.15 mm for the epidermis (instead of 0.2 mm), 1.5 mm for the dermis layer (as before), and 18.35 mm for the subcuteneous layer (instead of 18.3 mm). The \textbf{Combined} misspecification applies the offset shift of 0.1, uses the perturbed surrogate model $\tilde{f}_{LT}$ instead of the original $\hat{f}_{LT}$ for PPGen, and considers medium level noise for $\mathbf{x}_s$ and high noise for $\mathbf{x}_o$. The effect of these different types of misspecification is visualized in \autoref{fig:miss_viz}.

\begin{figure}
    \centering
    \includegraphics[width=\linewidth]{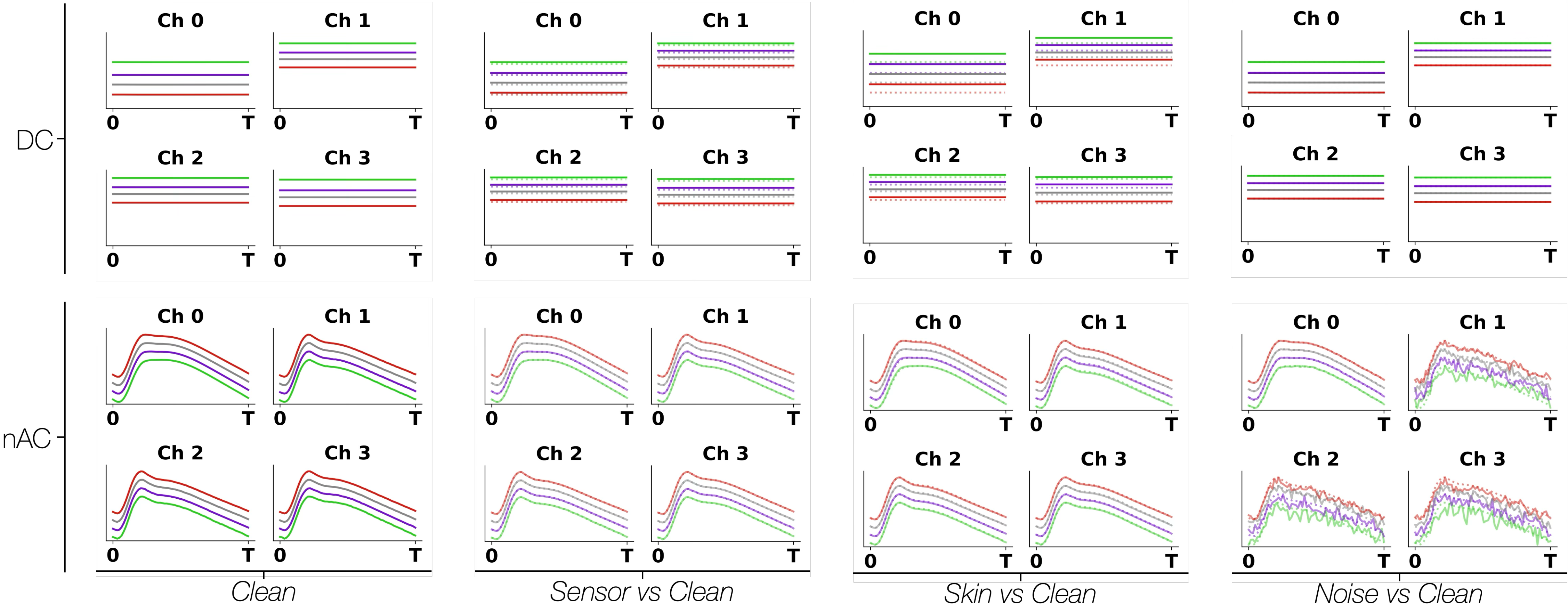}
    \caption{Vizualization of the misspecifications considered in our experiments against the same clean simulation (shown in dotted lines in the three right columns), for a randomly selected four-wavelength measurement. For the normalized AC we introduce an arbitrary vertical offset for the of vizualization. }
    \label{fig:miss_viz}
    \vspace{-1em}
\end{figure}

\begin{table*}[]
    \centering
    \begin{tabular}{ccc}
      \toprule
       \bfseries  Level & \ \bfseries White Gaussian Noise $\sigma$  & \bfseries Shot Noise $k_{shot}$ \\
       \midrule
       No noise & 0 & 0  \\
       Low noise & 1.0E-06 & 1.0E-07  \\
       Medium noise & 1.0E-05 & 1.0E-06  \\
       High noise & 1.0E-04 & 1.0E-05  \\
       Very high noise & 1.0E-03 & 1.0E-04  \\
       Extreme noise & 1.0E-02 & 1.0E-03  \\
       \bottomrule
    \end{tabular}
    \caption{Overview of considered noise levels. See \autoref{fig:noise_viz} for a visualization of noise. 
    }
    \label{tab:noise_levels}
\end{table*}

\section{HAI -- Algorithms and implementation}
\subsection{Algorithms}
\label{app:algs}
We provide a schematic of the HAI framework in \autoref{fig:diagram} and the sampling algorithm in \autoref{alg:PPGen}.

\begin{figure*}
    \centering
    \includegraphics[width=0.7\linewidth]{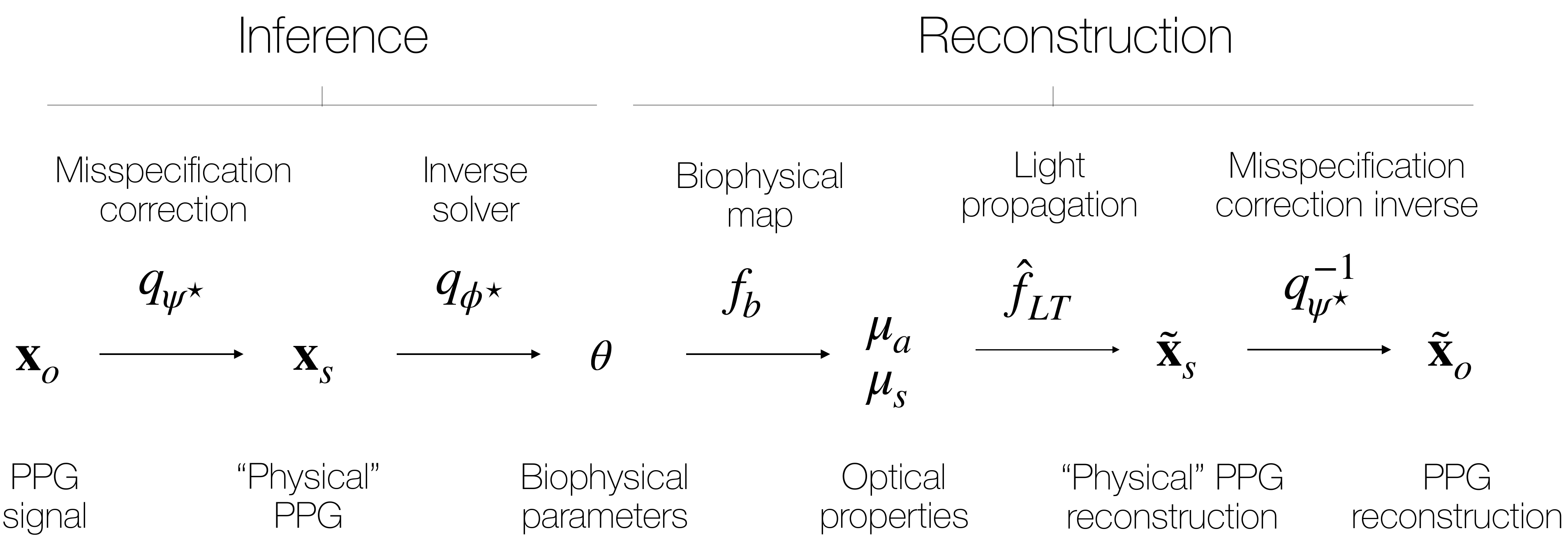}
    \caption{Schematic of the Hybrid Amortized Inference (HAI) framework.}
    \label{fig:diagram}
\end{figure*}

\begin{algorithm}[H]
\caption{Differentiable density evaluation and sampling for the PPG forward model.}
\label{alg:PPGen}
\DontPrintSemicolon

\SetKwFunction{EvaluateDensity}{EvaluateDensity}
\SetKwFunction{SampleFromModel}{SampleFromModel}
\BlankLine

\EvaluateDensity{$\theta,\, \mathbf{x}_s$}\;
Decompose $\theta$ into $\theta_s$ and $\{\theta_d^{\,t}\}_{t=1}^T$\;\\
$\ell \gets 0$\\ 
\For{$t=1,\dots,T$ \textit{ in parallel}}{
  \ForEach{$\lambda \in \Lambda$ \textit{ in parallel}}{
    $(\mu_a^{(t,\lambda)},\, \mu_s^{(\lambda)}) \gets f_b([\theta_s,\, \theta_d^{\,t}],\, \lambda)$\;
  }
  $\hat{\mathbf{x}}^{\,t}_s \gets \hat{f}_{\!LT}\big(\{\mu_a^{(t,\lambda)}\}_{\lambda \in \Lambda},\, \{\mu_s^{(\lambda)}\}_{\lambda \in \Lambda}\big)$\;\\
  $\ell \gets \ell + \log p\!\left(\mathbf{x}_{s}^t \mid \hat{\mathbf{x}}_{s}^t\right)$\;
}
\Return $\ell$\;

\BlankLine
\BlankLine
\SampleFromModel{$\Lambda,\, T$}\;\\
\textbf{(Prior sampling) Static:} \\$\theta_s \sim \text{Uniform}(\text{literature ranges})$\;\\
\textbf{(Prior sampling) Dynamic:} \\draw arterial pressure waves from a full-body hemodynamics simulator and map to dermis/subcutis blood-volume waveforms via Windkessel; form $\{\theta_d^{\,t}\}_{t=1}^T$ \;\\
$\theta \gets [\theta_s,\, \{\theta_d^{\,t}\}_{t=1}^T]$\;\\
\For{$t=1,\dots,T$\textit{ in parallel}}{
  \ForEach{$\lambda \in \Lambda$\textit{ in parallel}}{
    $(\mu_a^{(t,\lambda)},\, \mu_s^{(\lambda)}) \gets f_b([\theta_s,\, \theta_d^{\,t}],\, \lambda)$\;
    
  }
  $\hat{\mathbf{x}}^{\,t}_s \gets \hat{f}_{\!LT}\big(\{\mu_a^{(t,\lambda)}\}_{\lambda \in \Lambda},\, \{\mu_s^{(\lambda)}\}_{\lambda \in \Lambda}\big)$\;
}
Assemble $\hat{\mathbf{x}}_s \gets [\hat{\mathbf{x}}^{\,1}_s,\dots,\hat{\mathbf{x}}^{\,T}_s]$\;\\
$\mathbf{x}_{s} \sim p\!\left(\mathbf{x}_{s} \mid \hat{\mathbf{x}}_{s}\right)$\\
\Return $(\theta,\, \mathbf{x}_s)$\;

\end{algorithm}

\subsection{PPG features}
\label{app:ppg_features}
The pulsatile (AC) component of a PPG signal, which contains key morphological information, is typically much smaller in magnitude than the quasi-static (DC) baseline. Consequently, a standard reconstruction loss, such as the likelihood introduced in \autoref{sec:model}, can be dominated by the DC component, hindering the accurate reconstruction of the vital AC waveform. To address this, we decompose the observation $\mathbf{x}_o$ into its constituent parts. We define the DC component as the signal's temporal mean and the AC component as the mean-centered signal. We then compute the reconstruction loss using separate L2 norms on the AC and DC components, which is equivalent to assuming an isotropic Gaussian likelihood over these decomposed features. To further isolate the waveform's shape, we also compute a normalized AC (nAC) feature. These engineered features (AC, DC, and nAC) are also provided as augmented inputs to the encoder network $q_\phi$. Without this approach of augmenting both the loss function and the input features, capturing the subtle morphological changes in the PPG signal—and the physiological parameters that influence them—would be difficult due to the overwhelming amplitude of the DC component.

\section{Experimental details and extended results}
\label{app:exp_details}
Before setting the hyperparameters in our training workflow, we ran a hyperparameter search over most parameters. 

\subsection{Architectures and training details}

\paragraph{Normalization.}
We standardize the additional AC and normalized AC input features (see \autoref{app:ppg_features}) by computing the mean and standard deviation over 10 batches. The DC input features and biophysical parameters are standardized using the moments from the surrogate model training described in \autoref{app:surrogate_training}. 

\paragraph{Mapping parameter to specified ranges.}
Since the parameters $\theta$ are constrained to the ranges given in \autoref{tab:PSRs}, we need to map the real-valued outputs of the encoder to these ranges. For this we employ the cumulative distribution function (CDF) of the Gaussian distribution to map to $[0, 1]$. Afterwards we linearly transform the interval $[0, 1]$ to specified ranges for each parameter. 

\paragraph{Neural Posterior Estimator (NPE).}
The NPE, denoted as $q_\phi$, is implemented as a 1D U-Net, with the architecture described in \autoref{alg:1D_CNN_watch} for the four-wavelength setting and in \autoref{alg:1D_CNN_wide} for the wide-spectrum setting. In the four-wavelength setting, we flatten the receiver and wavelength dimensions (step 9), while we extract embeddings for the wide-spectrum setting (step 9-13). Afterwards the U-Net architecture processes the temporally embedded data and applies the decoding stage to predict dynamic biophysical parameters (two blood volume waveforms, $\Delta \text{BV}_2$ and $\Delta \text{BV}_3$). The intermediate representations from the U-Net's encoder are preserved for the subsequent static parameter inference head.

\begin{algorithm}
\caption{Neural network architecture for four-wavelength setting.}
\label{alg:1D_CNN_watch}
\begin{algorithmic}[1]
    
    \Function{ConvBlock}{$x, C_{in}, C_{out}$}
        \State $y \gets \text{Conv1d}(x, C_{in}, C_{out}, \text{kernel}=3, \text{padding}=1)$
        \State $y \gets \text{ReLU}(y)$
        \State $y \gets \text{Conv1d}(y, C_{out}, C_{out}, \text{kernel}=3, \text{padding}=1)$
        \State $y \gets \text{ReLU}(y)$
        \State \Return $y$
    \EndFunction
    \Statex 

    \Function{Forward}{$x \in \mathbb{R}^{N \times 4 \times 4\cdot 3 \times 64}$}
        \State $e_0 \gets \text{Flatten}(x, 1, 2)$ \Comment{Flatten wavelength and channel}
        \Statex \Comment{Encoder path}
        \State $e_1 \gets \text{ConvBlock}(e_0, 48, 32)$
        \State $e_2 \gets \text{ConvBlock}(\text{MaxPool1d}(e_1), 32, 32)$
        \State $e_3 \gets \text{ConvBlock}(\text{MaxPool1d}(e_2), 32, 16)$
        \State $e_4 \gets \text{ConvBlock}(\text{MaxPool1d}(e_3), 16, 8)$
        
        \Statex \Comment{Bottleneck}
        \State $b \gets \text{ConvBlock}(\text{MaxPool1d}(e_4), 8, 2 \cdot 8)$
        \Statex
        
        \Statex \Comment{Waveform decoder path with skip connections}
        \State $d_1 \gets \text{ConvTranspose1d}(b, 2 \cdot 8, 16, \text{kernel}=2, \text{stride}=2)$
        \State $d_1 \gets \text{Concatenate}(d_1, e_4)$ \Comment{Skip connection}
        \State $d_1 \gets \text{ConvBlock}(d_1, 8 + 16, 16)$
        
        \State $d_2 \gets \text{ConvTranspose1d}(d_1, 16, 32, \text{kernel}=2, \text{stride}=2)$
        \State $d_2 \gets \text{Concatenate}(d_2, e_3)$ \Comment{Skip connection}
        \State $d_2 \gets \text{ConvBlock}(d_2, 16 + 32, 32)$
        
        \State $d_3 \gets \text{ConvTranspose1d}(d_2, 32, 32, \text{kernel}=2, \text{stride}=2)$
        \State $d_3 \gets \text{Concatenate}(d_3, e_2)$ \Comment{Skip connection}
        \State $d_3 \gets \text{ConvBlock}(d_3, 32 + 32, 32)$
        
        \State $d_4 \gets \text{ConvTranspose1d}(d_3, 32, 32, \text{kernel}=2, \text{stride}=2)$
        \State $d_4 \gets \text{Concatenate}(d_4, e_1)$ \Comment{Skip connection}
        \State $d_4 \gets \text{ConvBlock}(d_4, 32 + 32, 32)$
        \State $d_5 \gets \text{GaussianSmoothing}(d_4, \text{sigma}=5.7, \text{kernel}=19)$
        
        \Statex \Comment{Inference head for dynamic parameters}
        \State $p_1 \gets \text{Conv1d}(d_5, 32, 32, \text{kernel}=5, \text{pad}=4, \text{dil}=2)$
        \State $p_2 \gets \text{Conv1d}(p_1, 32, 2, \text{kernel}=5, \text{pad}=2)$
        \Statex

        \Statex \Comment{Inference head for static parameters}
        \State $s_1 \gets \text{Concatenate}(e_1, e_2, e_3, e_4, b)$ \Comment{Skip connections}
        \State $s_2 \gets \text{MeanPool1D}(s_1)$ \Comment{Time averaging}
        \State $s_3 \gets \text{ReLU}(\text{LinearLayer}(s_2, 104, 600))$ 
        \State $s_4 \gets \text{ReLU}(\text{LinearLayer}(s_3, 600, 250))$ 
        \State $s_5 \gets \text{ReLU}(\text{LinearLayer}(s_4, 250, 125))$ 
        \State $s_6 \gets \text{LinearLayer}(s_5, 125, 9)$ 
        \Statex
        \Statex
        
    \EndFunction
\end{algorithmic}
\end{algorithm}

\begin{algorithm}
\caption{Neural network architecture for wide-spectrum setting.}
\label{alg:1D_CNN_wide}
\begin{algorithmic}[1]
    
    \Function{ConvBlock}{$x, C_{in}, C_{out}$}
        \State $y \gets \text{Conv1d}(x, C_{in}, C_{out}, \text{kernel}=3, \text{padding}=1)$
        \State $y \gets \text{ReLU}(y)$
        \State $y \gets \text{Conv1d}(y, C_{out}, C_{out}, \text{kernel}=3, \text{padding}=1)$
        \State $y \gets \text{ReLU}(y)$
        \State \Return $y$
    \EndFunction
    \Statex 

    \Function{Forward}{$x \in \mathbb{R}^{N \times 531 \times 4\cdot 3 \times 64}$}
        \Statex \Comment{Obtain embedding of wavelength and channel dimension}
        \State $w_1 \gets \text{Flatten}(x, 1, 2)$ \Comment{Flatten wavelength and channel}
        \State $w_2 \gets \text{ReLU}(\text{LinearLayer}(w_1, 531 \cdot 4 \cdot 3, 1000))$  \Comment{Linear layer separate across time}
        \State $w_3 \gets \text{ReLU}(\text{LinearLayer}(w_2, 1000, 1000))$  \Comment{Linear layer separate across time}
        \State $w_4 \gets \text{ReLU}(\text{LinearLayer}(w_3, 1000, 1000))$  \Comment{Linear layer separate across time}
        \State $w_5 \gets \text{LinearLayer}(w_3, 1000, 200)$  \Comment{Linear layer separate across time}
        \Statex \Comment{Encoder path}
        \State $e_1 \gets \text{ConvBlock}(w_5, 200, 32)$
        \State $e_2 \gets \text{ConvBlock}(\text{MaxPool1d}(e_1), 32, 64)$
        \State $e_3 \gets \text{ConvBlock}(\text{MaxPool1d}(e_2), 64, 64)$
        \State $e_4 \gets \text{ConvBlock}(\text{MaxPool1d}(e_3), 64, 64)$
        
        \Statex \Comment{Bottleneck}
        \State $b \gets \text{ConvBlock}(\text{MaxPool1d}(e_4), 64, 2 \cdot 64)$
        \Statex
        
        \Statex \Comment{Waveform decoder path with skip connections}
        \State $d_1 \gets \text{ConvTranspose1d}(b, 2 \cdot 64, 128, \text{kernel}=2, \text{stride}=2)$
        \State $d_1 \gets \text{Concatenate}(d_1, e_4)$ \Comment{Skip connection}
        \State $d_1 \gets \text{ConvBlock}(d_1, 64 + 128, 128)$
        
        \State $d_2 \gets \text{ConvTranspose1d}(d_1, 128, 64, \text{kernel}=2, \text{stride}=2)$
        \State $d_2 \gets \text{Concatenate}(d_2, e_3)$ \Comment{Skip connection}
        \State $d_2 \gets \text{ConvBlock}(d_2, 64 + 64, 64)$
        
        \State $d_3 \gets \text{ConvTranspose1d}(d_2, 64, 64, \text{kernel}=2, \text{stride}=2)$
        \State $d_3 \gets \text{Concatenate}(d_3, e_2)$ \Comment{Skip connection}
        \State $d_3 \gets \text{ConvBlock}(d_3, 64 + 64, 64)$
        
        \State $d_4 \gets \text{ConvTranspose1d}(d_3, 64, 32, \text{kernel}=2, \text{stride}=2)$
        \State $d_4 \gets \text{Concatenate}(d_4, e_1)$ \Comment{Skip connection}
        \State $d_4 \gets \text{ConvBlock}(d_4, 32 + 32, 32)$
        
        \Statex \Comment{Inference head for dynamic parameters}
        \State $p_1 \gets \text{Conv1d}(d_4, 32, 32, \text{kernel}=5, \text{pad}=4, \text{dil}=2)$
        \State $p_2 \gets \text{Conv1d}(p_1, 32, 2, \text{kernel}=5, \text{pad}=2)$
        \Statex

        \Statex \Comment{Inference head for static parameters}
        \State $s_1 \gets \text{Concatenate}(e_1, e_2, e_3, e_4, b)$ \Comment{Skip connections}
        \State $s_2 \gets \text{MeanPool1D}(s_1)$ \Comment{Time averaging}
        \State $s_3 \gets \text{ReLU}(\text{LinearLayer}(s_2, 352, 600))$ 
        \State $s_4 \gets \text{ReLU}(\text{LinearLayer}(s_3, 600, 250))$ 
        \State $s_5 \gets \text{ReLU}(\text{LinearLayer}(s_4, 250, 125))$ 
        \State $s_6 \gets \text{LinearLayer}(s_5, 125, 9)$ 
        \Statex
        
    \EndFunction
\end{algorithmic}
\end{algorithm}

\paragraph{Pretraining details for NPE (used in \autoref{results:noise}).}
We pretrain the four-wavelength NPE model for 2500 epochs and the wide-spectrum model for 2000 epochs. Each epoch consists of 100 iterations with a batch size of 200 samples. During pretraining we create synthetic data from PPGen on-the-fly, i.e. the model learns on unseen batches at each step of the training procedure. The loss function $\mathcal{L}_s(\phi)$ is implemented using the mean-squared-error (MSE). We use the AdamW optimizer \citep{loshchilov2017decoupled} with a learning rate of 0.0007 for the four-wavelength model and of 0.0002 for the wide-spectrum setting. Furthermore, we employ cosine annealing \citep{loshchilov2016sgdr} of the learning rate with $T_{max} =$ 1915 (1921) and $eta_{min} =$ \num{7.8e-5} (\num{4.9e-6}) (wide-spectrum setting in parentheses). Lastly, we regularize via weight decay with a coefficient of \num{2.8e-8} (\num{2.1e-9}). The best pretrained model is selected using the lowest validation loss, where is the validation set is drawn randomly from the PPGen simulator. 

\paragraph{Misspecification learning details (used in \autoref{results:misspec}).}
We train the misspecification correction layer $q_\psi$ for 500 epochs, while the NPE model is fixed with the pretrained model described above. Each epochs consists of 100 iterations with a batch size of 200 samples (before splitting off 20\% for validation). In contrast to the pretraining stage, we fix the dataset to mimic a fixed sized real-world dataset. The loss function is a combination of a MSE loss on DC features with a coefficient of 1, a MSE loss on AC with a coefficient of 0.1, and a MSE loss on normalized AC features with a coefficient of 0.1 (see \autoref{app:ppg_features} for a discussion of these common PPG features). To balance these different loss components more easily, we normalize each loss component with its value at initialization. This ensures that the coefficients at initialization directly reflect the weighting of each loss component. As before, we use the AdamW optimizer \citep{loshchilov2017decoupled} with a learning rate of 0.0001. We regularize via weight decay with a coefficient of \num{2.8e-8} (\num{2.1e-9}). The best model is selected using the lowest MAE loss for DC reconstruction on the validation set, with a train/val split of 80/20. 

\paragraph{Real-only baseline training details (used in \autoref{results:misspec}).}
For the Real-only baseline, we train both NPE model $q_\phi$ and misspecification correction model $q_\psi$ jointly using the same losses as during misspecification learning (1.0 for DC, 0.1 for AC, and 0.1 for normalized AC). In the four-wavelength setting we train the model for 2000 epochs and in the wide-spectrum setting for 200 epochs. As during misspecification learning, each epoch consists of 100 iterations with a batch size of 200 samples (before splitting off 20\% for validation). The other training hyperparameters are the same as in the pretraining setting for the NPE. The best model is selected using the lowest MAE loss for DC reconstruction on the validation set, with a train/val split of 80/20.

\paragraph{Sim-only baseline training details (used in \autoref{results:misspec}).}
This baseline consists of only training the NPE model. For this baseline, we can directly use the pretrained models.

\subsection{A detailed view of inference results using the four-wavelength PPG sensor}
\label{app:zoom_in_watch}

In \autoref{fig:zoom_in_static} we report inference results on static biophysical parameters using a four-wavelength PPG sensor without model misspecification, at medium noise level.

\begin{figure}[H]
    \centering
    \includegraphics[width=.5\linewidth]{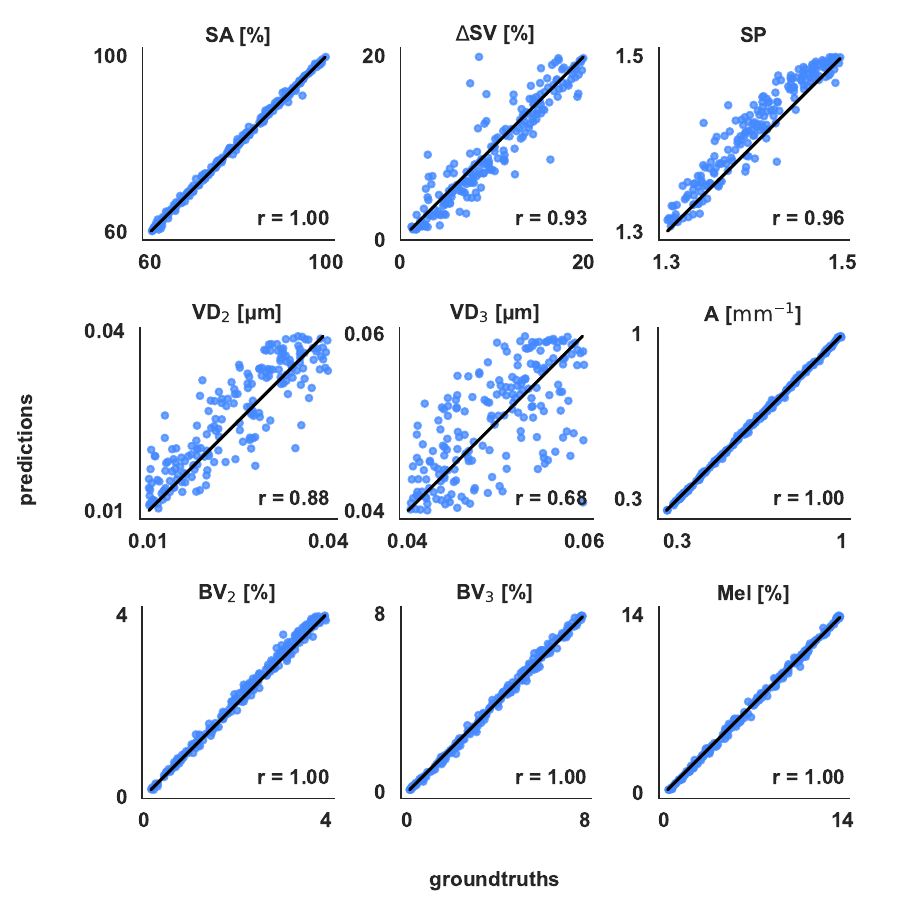}
    \caption{Inference of static biophysical parameters using a four-wavelength PPG sensor without misspecification and under medium noise level.}
    \label{fig:zoom_in_static}
\end{figure}

\subsection{Extended results for inference under misspecification}
\label{supp:results:misspec}
In \autoref{fig:misspec_watch_barplots} we provide a breakdown of inference results for each parameter under all misspecification settings for the four-wavelength sensor. In \autoref{fig:misspec_wide_barplots} we show the results for the wide-spectrum sensor. The aggregated results for the wide-spectrum sensor can be found in \autoref{tab:misspec_wide}.

\begin{table}[H]
    \floatconts
  {tab:results_misspec_wide}
  {\caption{Wide-spectrum setting: Parameter inference under several misspecifications, as measured by Pearson correlation and mean absolute percentage error~(MAPE) averaged over all biophysical parameters. For each result we report $\pm$ the standard deviation over multiple random seeds.}}
  {\label{tab:misspec_wide}}
  {\begin{tabular}{llll}
  \toprule
  \bfseries Misspec. & \bfseries Method & \bfseries Correlation  & \bfseries MAPE \\
  \midrule
  
  \multirow{ 3}{*}{\textit{None}} & HAI & $0.99 \pm 0.00$ & $1.0 \pm 0.0$  \\
       & Sim-only & $0.99 \pm 0.00$ & $1.0 \pm 0.0$ \\
       & Real-only & $0.72 \pm 0.01$ & $13.9 \pm 0.8$ \\
       \cmidrule(lr){1-4}
    
    \multirow{ 3}{*}{\textit{Noise}} & HAI & $0.97 \pm 0.00$ & $2.4 \pm 0.1$  \\
       & Sim-only & $0.97 \pm 0.01$ & $2.2 \pm 0.1$ \\
       & Real-only & $0.72 \pm 0.01$ & $14.0 \pm 0.9$ \\
       \cmidrule(lr){1-4}
    
    \multirow{ 3}{*}{\textit{Sensor}} & HAI & $0.99 \pm 0.00$ & $1.5 \pm 0.1$  \\
       & Sim-only & $0.94 \pm 0.00$ & $3.8 \pm 0.0$ \\
       & Real-only & $0.71 \pm 0.00$ & $15.7 \pm 0.8$ \\
       \cmidrule(lr){1-4}
    
    \multirow{ 3}{*}{\textit{Skin}} & HAI & $0.86 \pm 0.00$ & $9.6 \pm 0.3$  \\
       & Sim-only & $0.86 \pm 0.00$ & $9.0 \pm 1.2$ \\
       & Real-only & $0.72 \pm 0.01$ & $16.5 \pm 0.8$ \\
       \cmidrule(lr){1-4}
    
    \multirow{ 3}{*}{\textit{Combined}} & HAI & $0.84 \pm 0.00$ & $12.6 \pm 0.3$  \\
       & Sim-only & $0.84 \pm 0.00$ & $11.2 \pm 1.4$ \\
       & Real-only & $0.71 \pm 0.00$ & $15.2 \pm 0.2$ \\
    
  \bottomrule
  \end{tabular}}
\end{table}

\begin{figure*}
    \centering
    \includegraphics[width=0.6\textwidth]{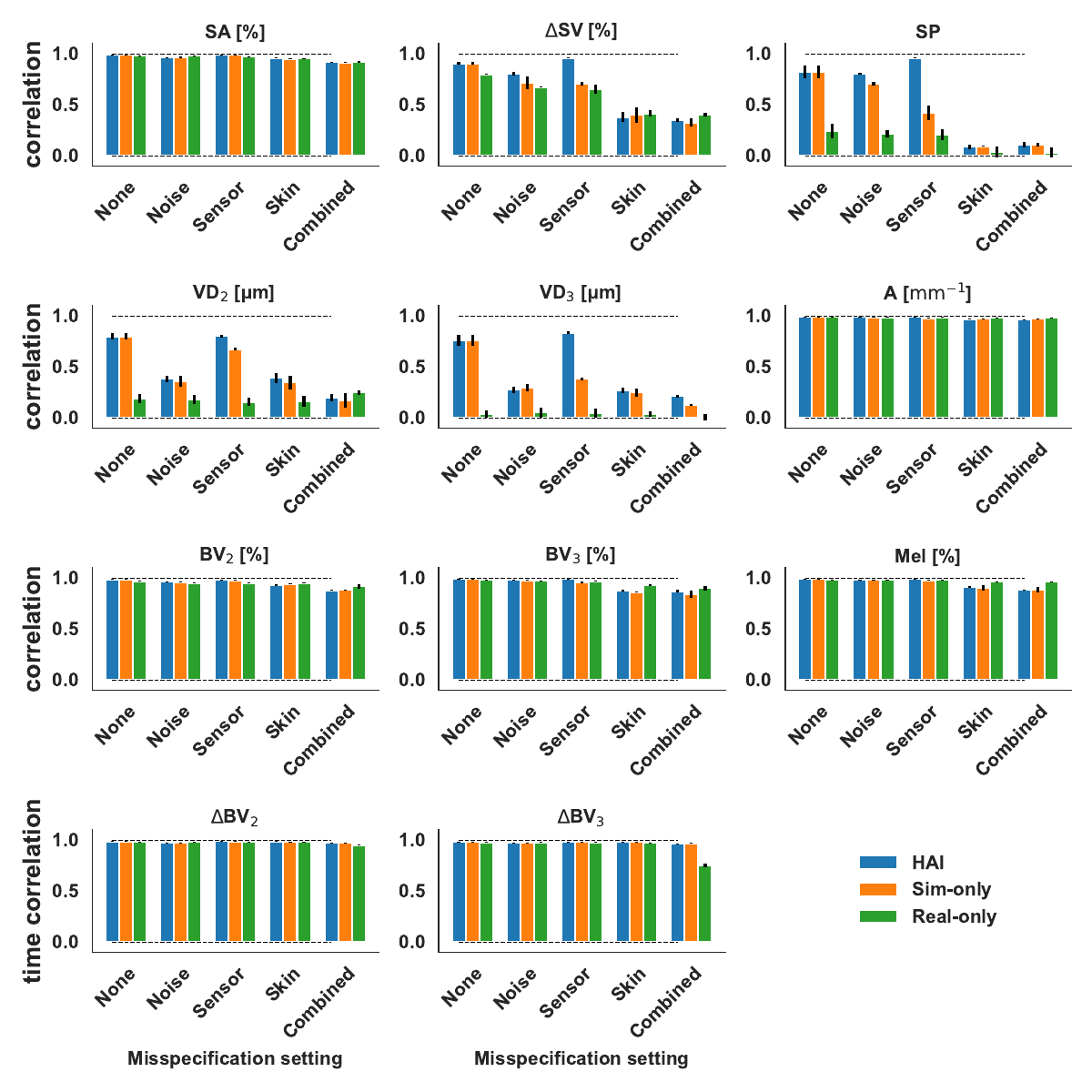}
    \hfill
    \includegraphics[width=0.6\textwidth]{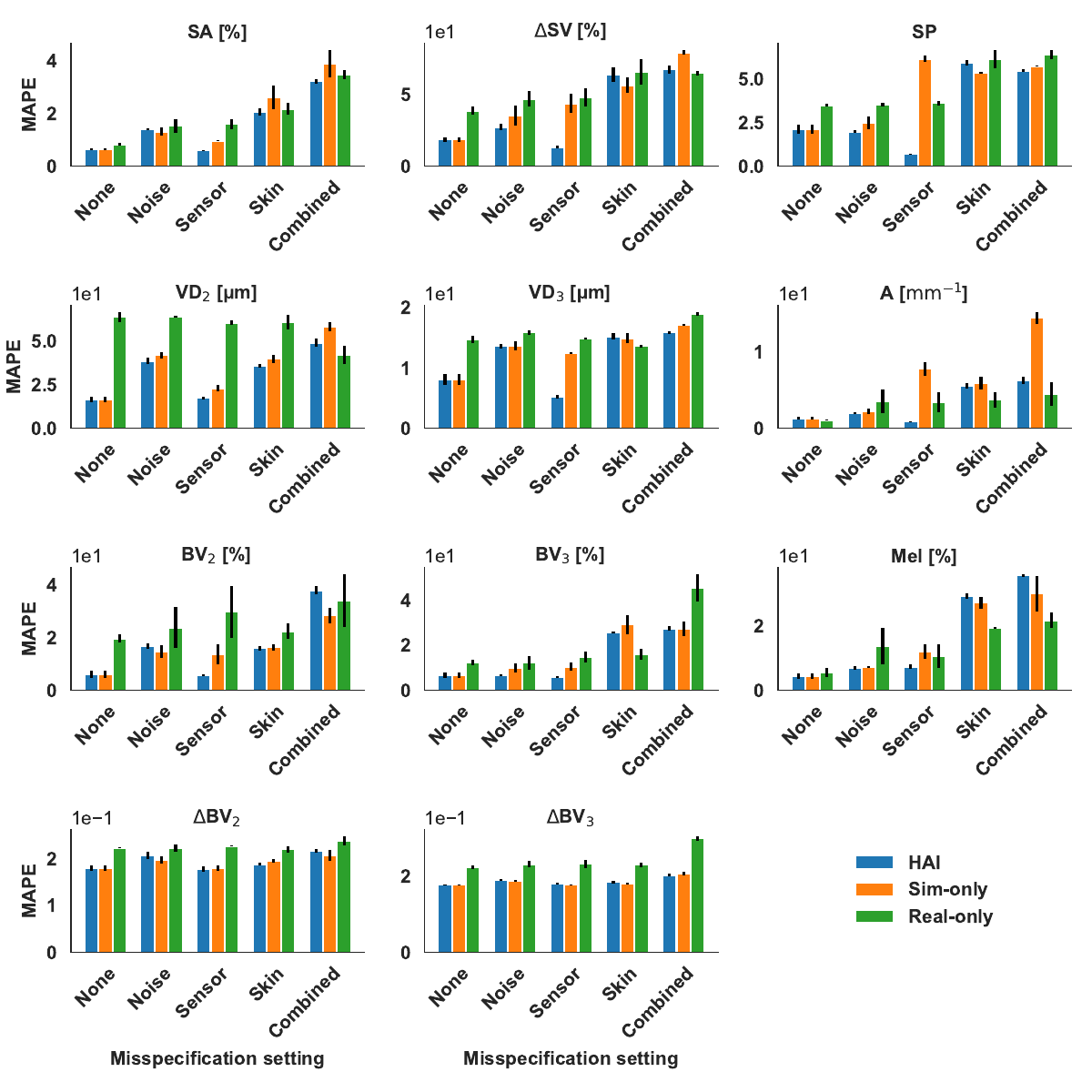}
    \caption{Four-wavelength PPG sensor: Tissue property inference under model misspecification. For each physiological parameter of interest, we report the Pearson correlation coefficient (upper panel) or mean-absolute percentage error (MAPE) (lower panel) between groundtruth and inferred parameters. For dynamic properties like systolic blood fractions, we report averages of across-time correlation coefficients.}
    \label{fig:misspec_watch_barplots}
\end{figure*}

\begin{figure*}
    \centering
    \includegraphics[width=0.6\textwidth]{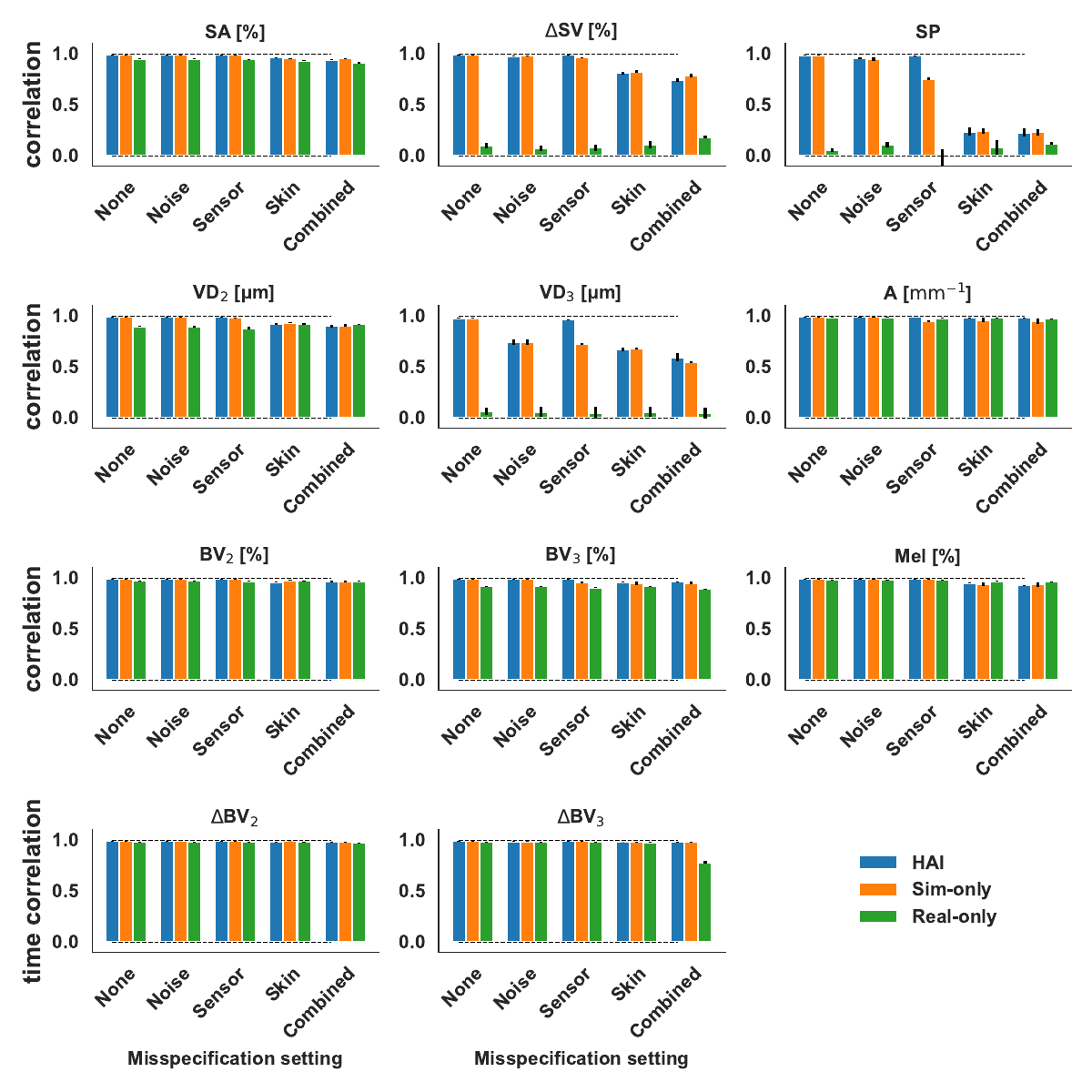}
    \hfill
    \includegraphics[width=0.6\textwidth]{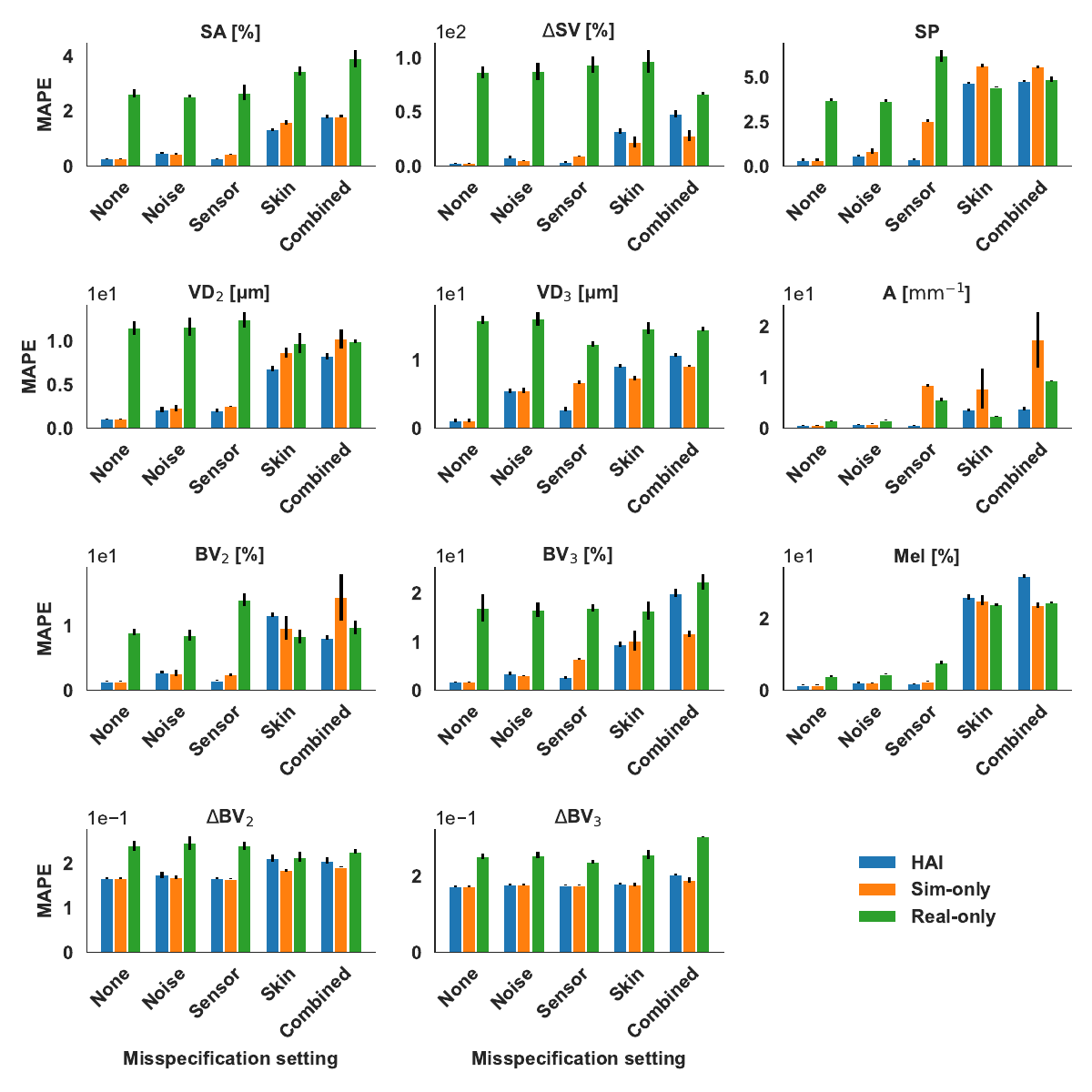}
    \caption{Wide-spectrum setting: Tissue property inference under model misspecification. For each physiological parameter of interest, we report the Pearson correlation coefficient (upper panel) or mean-absolute percentage error (MAPE) (lower panel) between groundtruth and inferred parameters. For dynamic properties like systolic blood fractions, we report averages of across-time correlation coefficients.}
    \label{fig:misspec_wide_barplots}
\end{figure*}

\end{document}